\documentclass{article}
\DeclareUnicodeCharacter{0301}{\'{e}}

\usepackage{microtype}
\usepackage{graphicx}
\usepackage{mathtools}
\usepackage{booktabs} % for professional tables
\usepackage{subcaption}
\usepackage{algorithm}
\usepackage{algpseudocode}

\usepackage{amsmath,amsfonts,bm}

\usepackage{hyperref}

\usepackage[nameinlink,capitalise]{cleveref}
\usepackage{etoc}
\etocframedstyle[1]{\textbf{\textsc{Table of Contents}}}

\usepackage[accepted]{icml2024}

\icmltitlerunning{Learning to Scale Logits for Temperature-Conditional GFlowNets}

\begin{document}

\twocolumn[
\icmltitle{Learning to Scale Logits for Temperature-Conditional GFlowNets}

\icmlsetsymbol{equal}{*}

\begin{icmlauthorlist}
\icmlauthor{Minsu Kim}{equal,to,kai}
\icmlauthor{Joohwan Ko}{equal,kai}
\icmlauthor{Taeyoung Yun}{equal,kai}
\icmlauthor{Dinghuai Zhang}{mila,mont}
\icmlauthor{Ling Pan}{hk}
\icmlauthor{Woo Chang Kim}{kai}
\icmlauthor{Jinkyoo Park}{kai}
\icmlauthor{Emmanuel Bengio}{recur}
\icmlauthor{Yoshua Bengio}{mila,mont,cifar}
\end{icmlauthorlist}

\icmlaffiliation{to}{Work performed while the author was at the Mila – Québec AI Institute}
\icmlaffiliation{kai}{Korea Advanced Institute of Science and Technology}
\icmlaffiliation{mila}{Mila – Québec AI Institute}
\icmlaffiliation{mont}{Université de Montréal}
\icmlaffiliation{hk}{Hong Kong University of Science and Technology}
\icmlaffiliation{recur}{Recursion}
\icmlaffiliation{cifar}{CIFAR}

\icmlcorrespondingauthor{Minsu Kim}{min-su@kaist.ac.kr}

\icmlkeywords{Machine Learning, ICML}

\vskip 0.3in
]

\printAffiliationsAndNotice{\icmlEqualContribution} % otherwise use the standard text.

\begin{abstract}
GFlowNets are probabilistic models that sequentially generate compositional structures through a stochastic policy. Among GFlowNets, temperature-conditional GFlowNets can introduce temperature-based controllability for exploration and exploitation. We propose \textit{Logit-scaling GFlowNets} (Logit-GFN), a novel architectural design that greatly accelerates the training of temperature-conditional GFlowNets. It is based on the idea that previously proposed approaches introduced numerical challenges in the deep network training, since different temperatures may give rise to very different gradient profiles as well as magnitudes of the policy's logits. We find that the challenge is greatly reduced if a learned function of the temperature is used to scale the policy's logits directly. Also, using Logit-GFN, GFlowNets can be improved by having better generalization capabilities in offline learning and mode discovery capabilities in online learning, which is empirically verified in various biological and chemical tasks. Our code is available at \url{https://github.com/dbsxodud-11/logit-gfn}
\end{abstract}

\section{Introduction}

Generative Flow Networks (GFlowNets) \cite{bengio2021flow} offer a training framework for learning generative policies that sequentially construct compositional objects to be sampled according to a given unnormalized probability mass or reward function. Whereas other generative models are trained to imitate a distribution implicitly specified by a training set, GFlowNets' target distribution is specified by a reward function seen as an unnormalized probability mass function. The primary advantage inherent to GFlowNets is their capacity to uncover a multitude of highly rewarded samples from a diverse set of modes \cite{bengio2021flow} of the given target distribution. This holds great significance in the context of scientific discovery, exemplified by domains such as drug discovery \cite{bengio2021flow,jain2022biological,jain2023multi}. 

Temperature-conditional GFlowNets are training frameworks for learning conditional generative models that generate samples proportional to a tempered reward function: $p(x|\beta) \propto R(x)^{\beta}$. In contrast to typical GFlowNets trained to match a single target distribution with fixed (inverse) temperature $\beta$, temperature-conditional GFlowNets learn a family of generative policies corresponding to reward functions raised to different powers. The major benefit of temperature-conditional GFlowNets is that it can adjust the generative policy based on the $\beta$, thereby allowing for the management of the exploration-exploitation trade-off. Furthermore, temperature-conditional GFlowNets effectively handle the training difficulties encountered with typical GFlowNets at lower temperatures (i.e., high $\beta$). These lower temperatures yield distributions that are extremely selective, posing a challenge for effective training using unconditional GFlowNets. Our expectation is that temperature-conditional GFlowNets, when trained at higher temperatures (low $\beta$ values), should be able to infer the distributions at lower temperatures (high $\beta$ values) more effectively. 

Temperature-conditional GFlowNets have already been introduced~\cite{zhang2022robust, zhou2023phylogfn} and have shown promising results for specific neural architectures, such as in the case of \textit{Topoformer} \cite{gagrani2022neural} for solving scheduling problems \cite{zhang2022robust}. Nonetheless, the empirical research on generic architectures remains limited. This limitation in research underscores the need for meticulous customization of temperature-conditional architectures to suit each specific task. Moreover, incorporating varying temperature parameters in the training process of neural networks has been observed to create challenges. These arise from the altered gradient profiles corresponding to each target temperature distribution, leading to training instability. For example, a higher temperature results in a peakier reward landscape, which poses challenges for stably training GFlowNets~\citep{malkin2022trajectory}. Thus, there is an urgent need for thorough empirical research into the behavior of temperature-conditional GFlowNets coupled with developing stable architectures to overcome these training challenges.

In this paper, we first suggest a new generic architecture design of temperature-conditional GFlowNets, called \textit{Logit-scaling GFlowNets} (Logit-GFN), to obtain a simple yet stable training framework. 
Our key idea is to integrate a direct pathway into the architecture. This pathway adjusts the policy's logits according to the parameter $\beta$, providing an effective inductive bias that adapts the target distribution's temperature. We hypothesize and verify experimentally that the Logit-GFN, which directly adjusts the logits in the sampling policy through temperature input, enhances the generalization and speeds up the training of temperature-conditional GFlowNets.

Additionally, we introduce an online discovery algorithm that utilizes Logit-GFN (i.e., $p(x|\beta)$) along with a dynamic control policy, $P_{\text{exp}}(\beta)$. The algorithm samples $\beta$ from the policy during online learning exploration. Then, we marginalize the Logit-GFN to train with decisions across multiple GFNs derived from different target distributions. This approach facilitates the discovery of new compositional structures within various tempered distributions, eliminating the need to train individuals GFN at different temperatures in every iteration. In our empirical study, we explore stationary distributions of $P_{\text{exp}}(\beta)$, including uniform, log uniform, normal, and dynamic distributions with simulated annealing, offering insights into the exploration-exploitation trade-off.

In our experimental results, the Logit-GFN architecture significantly enhances training stability, characterized by a smooth and rapid loss convergence. Moreover, it demonstrates improved offline generalization capability, showcasing its adeptness at generating novel and highly rewarded samples from fixed datasets. Our online learning with the Logit-GFN stands out with superior performance compared to GFN and alternative benchmarks, including well-established techniques in Reinforcement Learning (RL) \cite{schulman2017proximal, haarnoja2017reinforcement} and Markov Chain Monte Carlo (MCMC) methods \cite{xie2020mars}.

\section{Related Works}

Most generative AI approaches require a dataset to represent a target distribution to sample from, while GFlowNets are instead provided with an unnormalized probability mass, which we can consider as a reward function from an RL perspective. Methods for training and applying GFlowNets are rapidly evolving, demonstrating considerable progress across diverse domains, including causal discovery \cite{deleu2022bayesian,deleu2023joint}, combinatorial optimization \cite{zhang2022robust,zhang2023let}, biochemical discovery \cite{jain2022biological,jain2023gflownets}, reinforcement learning \cite{tiapkin2023generative} with world modeling~\cite{Pan2023StochasticGF}, large language model inference \cite{hu2023amortizing}, and diffusion-structured generative models \cite{Zhang2022UnifyingGM,lahlou2023theory, zhang2023diffusion}. The enhancements in GFlowNets are primarily attributed to advancements in training objectives, credit assignment techniques, and improved exploration strategies. The advent of temperature-conditional GFlowNets represents an exciting frontier in research, offering significant potential to elevate the performance and versatility of GFlowNets. In this context, we provide a comprehensive survey covering these directions.

\textbf{Training objective of GFlowNets.} GFlowNets, introduced by \citet{bengio2021flow}, open interesting new avenues. One core feature is the ability to have a greater diversity of modes of the target distribution compared with existing RL, variational methods, or MCMC. The other one is their off-policy training objectives: they can be trained on examples and trajectories from any distribution with full support, not necessarily from the distribution corresponding to their current parameters (because that would not allow sufficient exploration and diversity) or from the target distribution (for which one may not have samples). Recent notable advances include loss functions like trajectory balance (TB) \cite{malkin2022trajectory}, subtrajectory balance (SubTB) \cite{madan2023learning},
% aiming to tune the bias-variance trade-off finely.
and quantile matching (QM) \cite{zhang2023distributional}.

\textbf{Local credit assignment for GFlowNets.}  
Several approaches have been developed to enhance GFlowNets' training efficiency. Forward-Looking GFlowNet (FL-GFN) \cite{pan2023better} calculates intermediate energy from states, contributing to more stable training over longer trajectories. Building on this, the transition-based GFlowNet \cite{zhang2023let} specializes in combinatorial optimization. Additionally, \citet{jang2023learning} focus on decomposing energy into partial energies, achieving more effective local credit assignment compared with FL-GFN. Finally, \citet{falet2023deltaai} utilize the inductive bias of sparse graphical models to avoid having to ever evaluate full trajectories and terminal states, facilitating rapid learning in probabilistic graphical models.

\textbf{Better exploration for GFlowNets.} Another research focus is enhancing the exploration capabilities of GFlowNets. 
\citet{Pan2022GenerativeAF} propose a framework for introducing augmented flows into the flow network, which are represented by intrinsic curiosity-based incentives to encourage exploration in sparse environments.
\citet{rector2023thompson} apply Thompson sampling~\cite{osband2016deep} into the training of GFlowNets, while
\citet{shen2023towards} have advocated for prioritized replay training, a method that directs GFlowNets updates to concentrate on regions with higher rewards. Additionally, \citet{kim2023local} incorporates a back-and-forth refining process \cite{zhang2022generative} into its training algorithm. This integration of local search strategies significantly improves the quality of local exploration in GFlowNets, leading to the acquisition of higher-rewarded samples during training.

\begin{figure*}[t]
    \centering
    \includegraphics[width=0.9\textwidth]{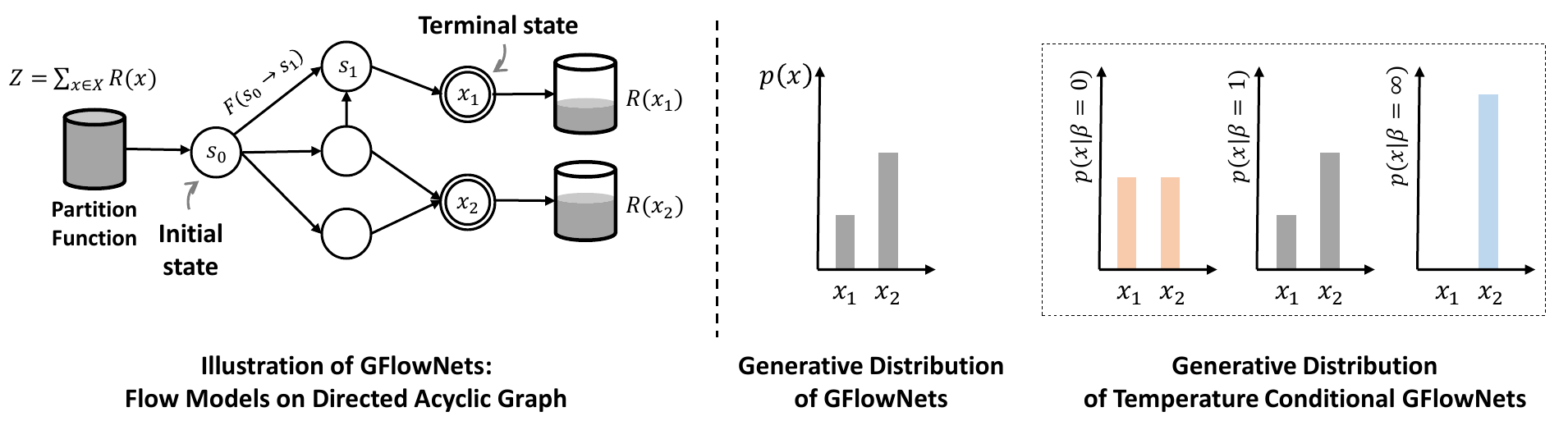}
    \caption{Illustration of GFlowNets and temperature conditional GFlowNets.}
    \label{fig:illustration}
    \vspace{-2ex}
\end{figure*}

\textbf{Temperature conditional GFlowNets.} Temperature conditioning, applied in combinatorial scheduling problems \cite{zhang2022robust}, uses a variable temperature factor to modulate the scheduling objective's smoothness. In the Topoformer architecture, \citet{gagrani2022neural} implement this through matrix multiplication with the temperature parameter in the initial linear layer. Similarly, \citet{zhou2023phylogfn} adopt temperature-conditional GFlowNets for phylogenetic inference, presenting a new approach to Bayesian variational inference with GFlowNets. Although these applications show promising results, there is a lack of concrete empirical evidence confirming the specific contribution of and issues with temperature conditioning. Recent studies in contrastive learning \citep{qiu2023not} and the development of a temperature prediction network for large foundation models \citep{qiu2024cool} have also investigated the issue of learning context-dependent temperature values. Our work, focusing on the impact of temperature conditioning through extensive empirical research, introduces a novel generic architecture aimed at stabilizing the training of temperature-conditional GFlowNets.

\begin{figure*}[t]
    \centering
    \includegraphics[width=0.95\textwidth]{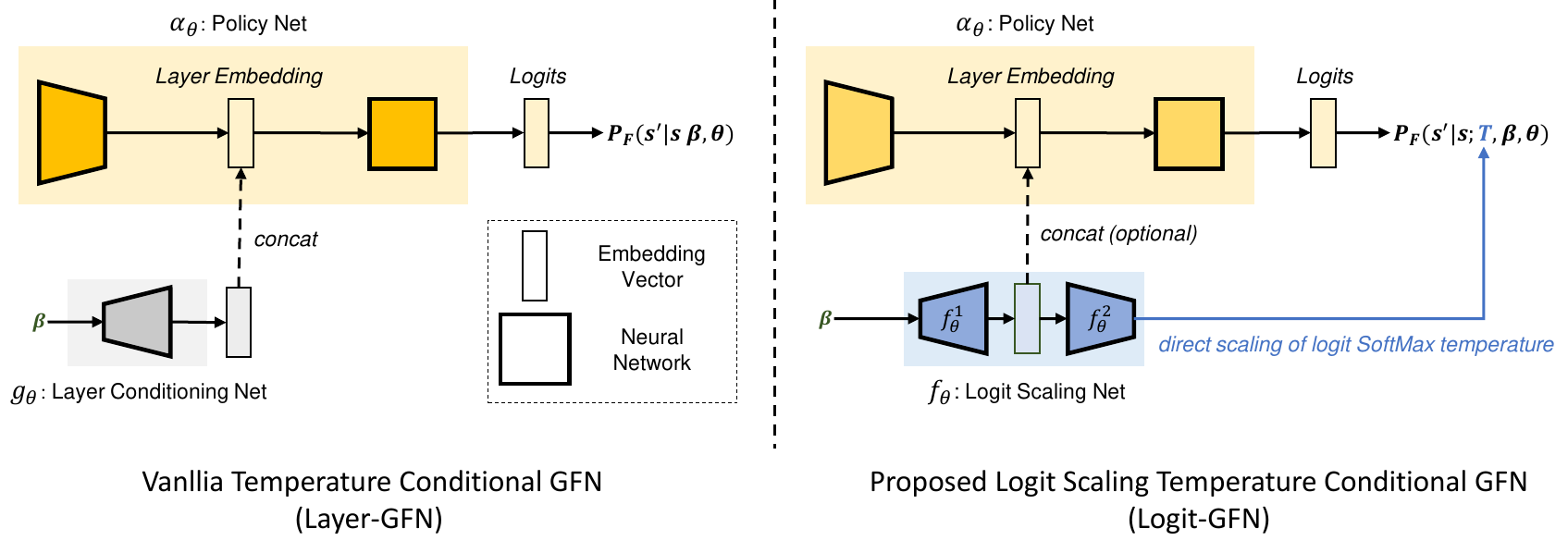}
    \caption{Architecture design of vanilla temperature conditional GFN and our Logit-GFN. The vanilla implementation of the temperature-conditional GFN integrates the embedding vector from $\beta$ by concatenating it with the layer embedding of the policy network. In contrast, the proposed method directly modulates the logit Softmax temperature.  }
    \label{fig:main}
    \vspace{-2ex}
\end{figure*}

\section{Preliminaries}

This section introduces GFlowNets and temperature conditional GFlowNets more formally; see \Cref{fig:illustration} for conceptual understanding and see \citet{bengio2023jmlr} for a full introduction. 
Generative flow networks (GFlowNets) constitute a class of deep generative models and a reinforcement learning (RL) methodology designed to sample compositional objects $x \in \mathcal{X}$, given a target distribution specified by an unnormalized probability mass or positive reward function. A generated object corresponds to the \textit{terminal state} in a Markov decision process starting from a unique initial state $s_0$. The GFlowNets policy sequentially adds an \textit{action} into a partial object (called the \textit{state}); the \textit{complete trajectory} $\tau = (s_0 \rightarrow \cdots \rightarrow s_n = x)$, which leads to compositional object $x$, is sequentially generated by the policy through a corresponding sequence of actions, each conditioned on the current state.

The space of all possible state sequences (trajectories) from the initial state to a terminal state is specified by a directed acyclic graph (DAG) that can incorporate domain-specific constraints (about which action is allowed in any state). It is noteworthy that, in opposition to a tree structure, the more general DAG structure offers a pivotal advantage by enabling the modeling of numerous potential action sequences that converge to identical states. This is different from soft Q-learning and entropy-regularized RL~\citep{haarnoja2017reinforcement,haarnoja2018soft,buesing2020approximate}, which are closely related to GFlowNets but may misbehave in the DAG setting where there are multiple ways of landing in the same terminal state~\citep{bengio2021flow}.

\textbf{Definition for flows.} The \textit{trajectory flow}, denoted by $F(\tau)$, is a non-negative function that maps complete trajectories to unnormalized probabilities, representing the flow of probability from the initial state to a terminal state along that trajectory, with the idea that the total flow into a terminal state $x$ should match the reward function value at $x$, $R(x)$.

This flow is further dissected into \textit{state flows} and \textit{edge flows}. The state flow, represented by $F(s)$, is the sum of $F(\tau)$ over all trajectories $\tau$ containing state $s$. Similarly, $F(s \rightarrow s')$ is the sum of flows of trajectories with a transition from state $s$ to state $s'$.

\textbf{Relationship between flow and policy.} The \textit{forward policy}, $P_F({s_{t+1}|s_t})$, quantifies the probability of transitioning from a state to any of its child states, while the \textit{backward policy}, $P_B({s_{t}|s_{t+1}})$, captures the probability of transitioning from a state to one of its parent states. When $F$ is a \textit{Markovian flow}, the forward policy and backward policy can derived as follows: $P_F(s'|s) = F(s \rightarrow s')/F(s)$ and $P_B(s|s') = F(s \rightarrow s')/F(s')$. 

\textbf{Marginal distribution of GFlowNets.} Eventually, we define the key concept, a marginal distribution, denoted as $P_{F}^{\top}(x):= \sum_{\tau \rightarrow x}P_F(\tau)$, which aggregates probabilities of trajectories terminating at a specific state $x$. The learning problem of GFlowNet is approximately achieving the following conditions: 
{%
\setlength{\belowdisplayskip}{1ex} \setlength{\belowdisplayshortskip}{1ex}
\setlength{\abovedisplayskip}{1ex} \setlength{\abovedisplayshortskip}{1ex}
 \begin{equation}
   P_{F}^{\top}(x) = \sum_{\tau \rightarrow x}P_F(\tau)   = \frac{R(x)}{\sum_{x \in \mathcal{X}}R(x)},  \label{eq:gfn}
 \end{equation}
 }%
where \(R(x)>0\) refers to a reward for a terminal state \(x\).

\textbf{Learning objective of GFlowNets.} \Cref{eq:gfn} can be satisfied by minimizing a GFlowNet training objective. The most commonly used objective is trajectory balance (TB) \cite{malkin2022gflownets}. Given the GFlowNet with parameters \(\theta\), the core principle of TB mandates that the flow of a trajectory computed forward (from the initial state to a terminal state) must match the flow computed backward (from a terminal state to the initial state):
{%
\setlength{\belowdisplayskip}{1ex} \setlength{\belowdisplayshortskip}{1ex}
\setlength{\abovedisplayskip}{1ex} \setlength{\abovedisplayshortskip}{1ex}
\begin{align}
    Z_{\theta}\prod_{t=1}^n P_F(s_{t+1}|s_t;\theta) = R(x)\prod_{t=1}^n P_B(s_{t}|s_{t+1};\theta).
\end{align}
}%
Here, $Z_{\theta}$ estimates the flow through the initial state $s_0$ and it also estimates the partition function, the sum of all trajectory flows: $Z_\theta = \sum_{\tau \in \mathcal{T}}F(\tau) = \sum_{x \in \mathcal{X}}R(x)$ when the constraint is satisfied, i.e., when the corresponding mismatch loss is minimized.

Other loss functions have been proposed to satisfy \Cref{eq:gfn}, such as sub-trajectory balance \cite{madan2023learning}, guided-trajectory balance \cite{shen2023towards} and detailed balance \cite{bengio2023jmlr}.

\subsection{Temperature conditional GFlowNets}

The goal of the temperature-conditional GFlowNets is to train conditional generative models proportional to a tempered reward function, $p(x\vert\beta)\propto R(x)^{\beta}$; see \Cref{fig:illustration} for illustration. The models have to take the (inverse) temperature $\beta$ as an additional input to represent a temperature-conditional distribution.
A conventional approach for constructing a conditional model involves concatenating the conditioning values directly into model layers \citep{song2020score, ho2020denoising, zhang2022robust}. We denote this approach as {\em layer-conditioned} GFlowNet (Layer-GFN) that integrates temperature embeddings directly into the model parameterized by $\theta$. The temperature embedding is a fixed or learned function $g_\theta : \mathbb{R}\rightarrow \mathbb{R}^{d}$, where \(d\) denotes the dimension of the temperature embedding.

\textbf{Layer-GFN} concatenates the output of the temperature embedding function, $g_\theta(\cdot)$ with the output of the state-input embedding module in order to produce logits for the forward and backward policies (and flows if desired). For example, the forward policy is parameterized as follows:
{%
\setlength{\belowdisplayskip}{1ex} \setlength{\belowdisplayshortskip}{1ex}
\setlength{\abovedisplayskip}{1ex} \setlength{\abovedisplayshortskip}{1ex}
\begin{align}
P_F^{\mathrm{Layer}}(s'|s;\beta,\theta):= \frac{\exp(\alpha_{\theta}(s,s',g_{\theta}(\beta)))}{\sum_{s'' \in Ch(s)} \exp(\alpha_{\theta}(s,s'',g_{\theta}(\beta)))}   
\end{align}
}%
\textbf{Limitations} 
While the Layer-GFN is a useful method for constructing temperature-conditional GFlowNets, our experiments also suggest that training temperature-conditional GFlowNets is numerically more difficult than training unconditional GFlowNets for each temperature. This motivated the variants of logit scaling proposed below.

\section{Methodology}

We propose Logit-GFN, a novel temperature-conditional GFlowNet that addresses the numerical challenges identified in previous versions of temperature-conditional GFlowNets. 
Specifically we introduce \textit{logit-scaling} trick for Logit-GFN that directly adjusts the output softmax temperature ($T$) based on the specified input inverse temperature $\beta$. This provides a direct pathway for controlling the softmax temperatures of the forward policy, creating an effective inductive bias for adapting to changes in the target distribution's temperature. This, in turn, ensures more stable training. We verify the effectiveness of the proposed architecture in training stability and offline generalization capability. Also, we provide a novel online learning algorithm by leveraging the Logit-GFNs, which can effectively discover novel combinatorial structures (i.e., modes) in scientific discovery tasks. See the right side of \Cref{fig:main} for the overall architecture of Logit-GFN. 

\subsection{Logit scaling}

The objective of the method aligns with that of layer-conditioning: to facilitate the training of temperature-conditional GFlowNets $p(x|\beta) \propto R(x)^{\beta}$ over varying inverse temperatures $\beta$. Logit-scaling trick use a simple skip connection to adjust the softmax temperature $T$ of logits 
of $P_F$ as a direct function of $\beta$. More specifically, logit-scaling into policy net $\alpha_{\theta}$ can be defined as follows: 
{%
\setlength{\belowdisplayskip}{1.5ex} \setlength{\belowdisplayshortskip}{1.5ex}
\setlength{\abovedisplayskip}{1.5ex} \setlength{\abovedisplayshortskip}{1.5ex}
\begin{align}
 P_F(s'|s;\beta,\theta):= \frac{\exp\left(\alpha_{\theta}\left(s,s'\right)/f_{\theta}\left(\beta\right)\right)}{\sum_{s'' \in Ch(s)} \exp(\alpha_{\theta}(s,s'')/f_{\theta}\left(\beta\right))},
\end{align}
}%
where $\alpha_{\theta}:S\times S\rightarrow\mathbb{R}$ now becomes neural net that is independent of $\beta$ and  $f_{\theta}: \mathbb{R} \rightarrow \mathbb{R}$ is the \textit{logit scaling} net, which transforms the inverse temperature \(\beta\) into a softmax  temperature, \(T=f_{\theta}\left(\beta\right)\). Note that the logit-scaling method is agnostic to the policy network; the policy network can take any form, including a layer-conditioning network.

For a detailed parameterization of logit-scaling, the logit-scaling network, \(f_{\theta}\), consists of an encoder and a decoder. The encoder, $f^1_{\theta}: \mathbb{R} \rightarrow \mathbb{R}^D$, maps a scalar input to a $D$-dimensional embedding vector. Conversely, the decoder, $f^2_{\theta}: \mathbb{R}^D \rightarrow \mathbb{R}$, converts this embedding vector back into a scalar. The overall transformation is represented by the composition $f = f^1_{\theta} \circ f^2_{\theta}$, seamlessly integrating the encoder and decoder functionalities. Given the logit scaling net, the softmax temperature $T=f_{\theta}\left(\beta\right)$ determines the confidence level of $P_F$ as a lower $T$ makes a sharper decision, and a higher $T$ gives a smoother decision. By adjusting the scalar value $T$ as a function of $\beta$, we can adjust the output generative distribution easily without heavy parameterization. The training objective makes the logit scaling net, \(f_{\theta}\), adjust the target temperature-conditional forward policy towards matching the tempered reward function.  

\textbf{Layer-conditioning with logit scaling.}  Utilizing the logit scaling network, \(f_{\theta}\), the Logit-GFN provides the flexibility to incorporate layer-conditioning using a simple yet intuitive technique. We start by taking the output of the encoder, \(f_{\theta}^1\left(\beta\right)\), which serves as a latent temperature embedding vector. This vector can then be seamlessly concatenated with the layer embedding of the policy net, \(\alpha_{\theta}\), as depicted in \Cref{fig:main}. Since the expressive power of \textit{logit-scaling} is limited when \(\alpha_{\theta}\) is used unconditionally with respect to \(\beta\), integrating it with the \textit{layer-conditioning} method can be a viable option to achieve full expressive power for temperature conditioning. We explore it further in the \cref{app:ablation_logit+layer}.

\subsection{Training objective}

The training procedure for our experiments is based on the trajectory balance (TB) loss, so we aim to minimize the TB loss given a training replay buffer or dataset $\mathcal{D}$ similar to prior work \cite{shen2023towards}. The difference is that we train the GFlowNets with multiple values of $\beta \sim P_{\text{train}}(\beta)$: 
{
\setlength{\belowdisplayskip}{1.5ex} \setlength{\belowdisplayshortskip}{1.5ex}
\setlength{\abovedisplayskip}{1.5ex} \setlength{\abovedisplayshortskip}{1.5ex}
\begin{align}
\label{eq:loss}
    \mathcal{L}(\theta;\mathcal{D}) 
    = \mathbb{E}_{P_{\text{train}}(\beta)}&\mathbb{E}_{P_{\mathcal{D}}(\tau)}\left[\left(\log\frac{Z_{\theta}(\beta)\prod_{t=1}^n P_F(\cdot)}{R(x)^{\beta}\prod_{t=1}^nP_B(\cdot)}\right)^2\right].
    \\
    \mathrm{where} \quad
    &P_F(\cdot) = P_F(s_t|s_{t-1};\beta,\theta) \nonumber
    \\
    &P_B(\cdot) = P_B(s_{t-1}|s_{t};\beta,\theta) \nonumber
\end{align}
}

When considering the parameterization of deep neural networks (DNNs) using $\theta$, a key implementation feature %distinction between our approach and previous techniques 
lies in the conditioning of the partition function $Z$ on the inverse temperature $\beta$, i.e., we write $Z_{\theta}(\beta)$ as a learned function rather than a learned constant.   Additionally, the conditional dependencies of $P_F$ and $P_B$ on \(\beta\) are established by the DNN $f_{\theta}$. This architecture necessitates the incorporation of two auxiliary DNNs, namely $Z_{\theta}$ and $f_{\theta}$; however, it's worth noting that these mappings operate from a scalar to a scalar, mitigating the need for an excessive number of parameters. This training objective and the mathematical theory for conditional GFlowNets was originally proposed by \cite{bengio2023jmlr}.

\subsection{Online discovery algorithm with Logit-GFN}

In scientific discovery (e.g., molecule optimization), we aim to discover a set of diverse candidate objects $x \in \mathcal{X}$ with high reward $R(x)$, e.g., molecules with high binding affinity to some protein. What we care about is not only the top rewards among these candidates but also their diversity and the number of modes (a local peak in which the reward is above a certain threshold) ~\citep{jain2022biological}, due to the uncertainty and imperfections of the reward functions.

Typically, training of GFlowNets involves an exploratory policy that generates trajectories that can be put in a prioritized replay buffer and are then used to perform gradient updates on the GFlowNets parameters. 
We can significantly enhance the exploratory phase of GFlowNets by querying multiple values of $\beta$ with temperature-conditional GFlowNets when forming samples for the replay buffer. This enables the model to generate a diverse set of candidates sampled from various generative distributions: 
{%
\setlength{\belowdisplayskip}{1.5ex} \setlength{\belowdisplayshortskip}{1.5ex}
\setlength{\abovedisplayskip}{1.5ex} \setlength{\abovedisplayshortskip}{1.5ex}
\begin{align}
    \mathcal{D} 
    &\gets \mathcal{D} \cup \{\tau_1,\ldots,\tau_M\}\\ \quad \tau_1,\ldots,\tau_M 
    &\sim \int_{\beta}P_F(\tau|\beta)dP_{\text{exp}}(\beta).
\end{align}
}%
The dynamic control policy $P_{\text{exp}}(\beta)$ controls the range of $\beta$ for exploration, and it may be chosen differently from $P_{\rm train}(\beta)$ in Eq.~\ref{eq:loss}. We provide several empirical observations for different $P_{\text{exp}}$ in \cref{beta_exploration} and \cref{app:query_distribution}. See \Cref{alg:tempgfn}
for full pseudocode.

\section{Experiments}

We present experimental results on 4 biochemical tasks: QM9, sEH, TFBind8, and RNA-binding. Following recent work \cite{shen2023towards}, we formulate the problem as a sequence prepend/append MDP, where the actions add a token either to the leftmost or rightmost end of a partial sequence or molecule. This setting results in multiple trajectories $\tau$ for each sample $x$, where the DAG structure is crucial for exploring the trajectory space. We describe the tasks below. 

\textbf{QM9} This task requires generating a small molecule graph. We build a graph using 12 building blocks with 2 stems, and each molecule consists of 5 blocks. Our objective is to maximize the HOMO-LUMO gap \citep{zhang2020molecular}.

\textbf{sEH} This task requires generating a small molecule graph. We build a graph using 18 building blocks with 2 stems, and each molecule consists of 6 blocks. Our objective is to maximize binding affinity \citep{bengio2021flow}.

\textbf{TFBind8} This task requires generating a string of 8 nucleotides. The objective is to maximize the DNA binding affinity to a human transcription factor, SIX6, from \citep{trabucco2022design}. 

\textbf{RNA-Binding} This task requires generating a string of 14 nucleobases. The objective is to maximize the binding affinity to the target transcription factor. We use L14-RNA1 as the target factor, introduced by \citet{sinai2020adalead}.

\subsection{Evaluation of training stability}\label{sec:training_stability}

We first assess the training stability of temperature-conditional GFlowNets. 

To evaluate stability, we compute the TB loss using samples generated by the forward policy under extremely high $\beta$ conditions ($\beta=5,000$). During off-policy training (with exploration), we sample multiple $\beta$ values from a range of relatively low temperatures,  ($\beta\sim U^{[10, 50]})$.  For unconditional GFlowNets, we implement an unconditional policy that is independent of the temperature and put the original reward to the power of 5,000. For additional details on the experimental setup, please see \Cref{app:training_stability}.

\begin{figure}[t]
    \centering
    \includegraphics[width=\columnwidth]{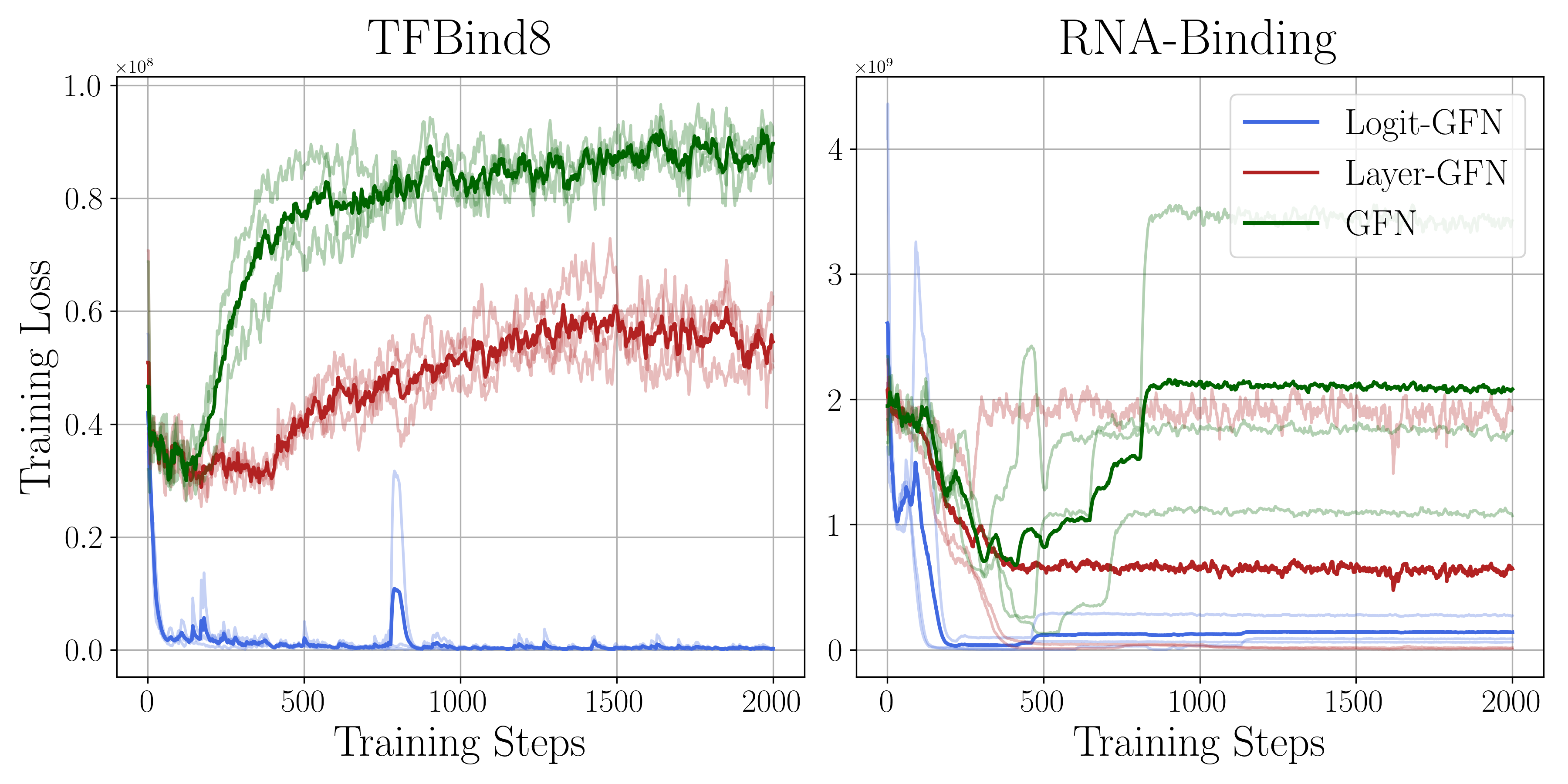}
    \vspace{-20pt}
    \caption{Loss of Temperature-conditional GFlowNets and unconditional GFlowNets as a function of a number of training steps on the TFBind8 and RNA-Binding tasks. Logit-GFN yields more stable training curves and converges faster. We draw curves with three different random seeds and highlight the mean over seeds.}
    \label{fig:loss_curve}
    \vspace{-10pt}
\end{figure}

\Cref{fig:loss_curve} illustrates the loss curves for both temperature-conditional and unconditional GFlowNets in TFbind8 and RNA-binding tasks. It is evident that training unconditional GFlowNets with high $\beta$ values leads to extreme instability. Although Layer-GFN alleviates this issue to some extent, it still exhibits considerable instability. Our novel approach, Logit-GFN, stands out by maintaining stable loss levels even when operating with high, previously unseen inverse temperatures.
This underscores the training stability of our proposed logit scaling in the development of temperature-conditional GFlowNets.

\begin{figure*}[t]
    \centering
    \includegraphics[width=0.95\textwidth]{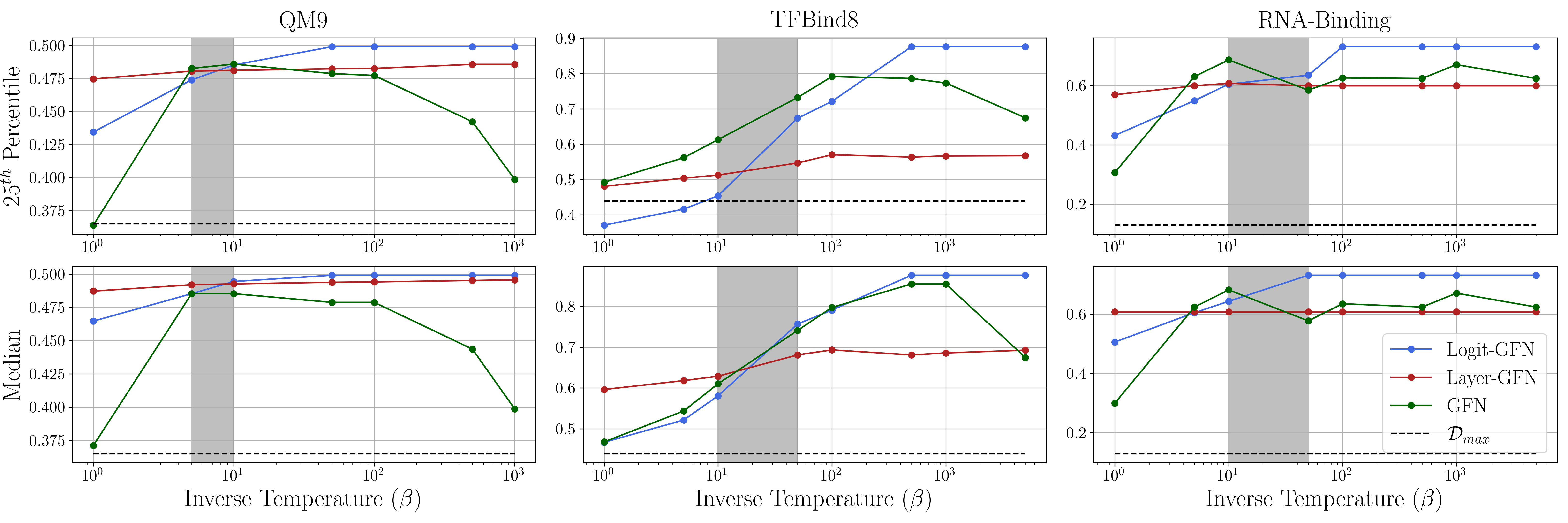}
    \vspace{-10pt}
    \caption{Performance of Temperature-conditional GFlowNets and unconditional GFlowNet in offline generalization. Shaded regions denote the temperature range used in training. Logit-GFN generates high-rewarding samples that surpass the offline datasets when conditioned on high $\beta$ values.}
    \label{fig:offline_generalization}
\end{figure*}

\begin{figure*}[t]
    \centering
    \includegraphics[width=0.95\textwidth]{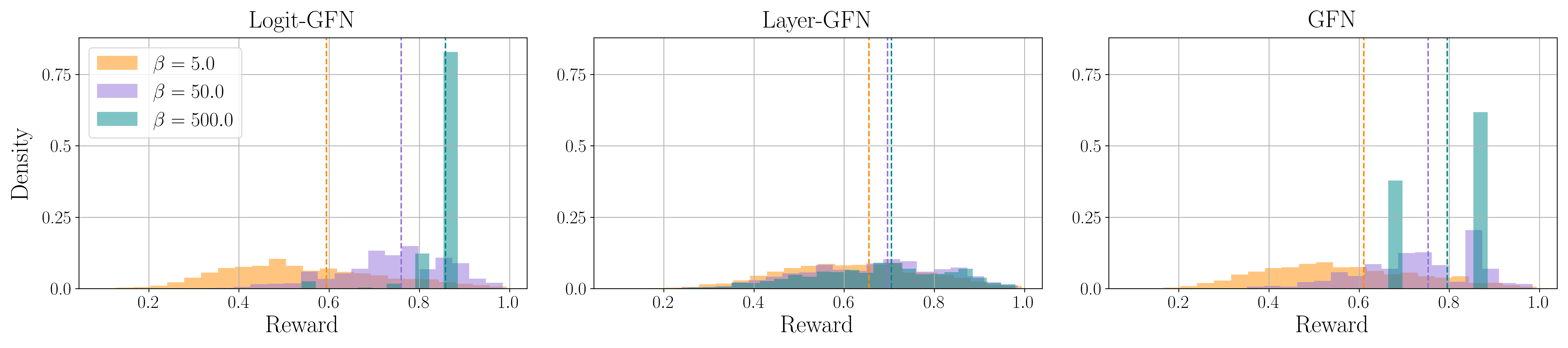}
    \vspace{-15pt}
    \caption{Reward distribution of samples from Temperature-conditional GFlowNets and unconditional GFlowNet in offline generalization. Logit-GFN dynamically shifts its reward distribution towards a high-reward region when conditioned on high $\beta$ values.}
    \label{fig:tfbind8_histogram}
\end{figure*}
\subsection{Evaluation of offline generalization}
In this section, we examine the controllability of temperature-conditional GFlowNets, specifically the ability to induce a relationship: $p(x\vert\beta)\propto R(x)^{\beta}$. To verify this, we prepare an offline dataset $\mathcal{D}=\{(x_i, R(x_i))\}_{i=1}^{N}$, similar to offline model-based optimization \citep{trabucco2022design}. 

We train temperature-conditional GFlowNets with relatively low values of $\beta$ and generate samples by querying high $\beta$ to verify that we can find high-scoring samples that surpass the offline dataset via out-of-distribution generalization. We also train unconditional GFlowNets trained with a high, fixed $\beta$ independently for comparison. To evaluate performance, we use $25^{th}$ percentile and median reward of samples generated from trained models, which can be used as an indicator of how trained policies adaptively respond to varying temperatures. Note that we train a single temperature-conditional model and investigate its generalization performance by querying with different $\beta$ values. For details on the experiment setting, please refer to \Cref{app:offline_generalization}.

\Cref{fig:offline_generalization} shows the offline generalization performance of GFlowNets in three biochemical tasks: QM9, TFBind8, and RNA-binding. The shaded region indicates the training range for temperature-conditional GFlowNets. We refer to it as in-distribution and consider querying outside of the shaded region as an out-of-distribution generalization. The figure demonstrates that Logit-GFN exhibits powerful generalization performance in high $\beta$ and is able to generate high-rewarding samples that surpass the offline dataset, corresponding to the queried values. However, training unconditional GFlowNets with high $\beta$ results in degraded performance despite their specialization to the fixed high $\beta$. While Layer-GFN often shows promising results on high $\beta$, it seems that Layer-GFN does not adaptively respond to different temperatures and tends to return relatively high-reward samples.
% the distribution of rewards remains relatively unchanged across varying $\beta$ values.

\Cref{fig:tfbind8_histogram} demonstrates Logit-GFN's ability in accurately capturing the reward distribution of samples across various temperatures. While Layer-GFN is unable to effectively shift its reward distribution according to $\beta$, the figure distinctly highlights that Logit-GFN dynamically adapts to different $\beta$ values. Unconditional GFlowNets individually trained with their respective $\beta$ values converge to generate sub-optimal samples, necessitating temperature-conditional GFlowNets, which can generalize across temperatures.

\begin{figure}[h]
    \centering
    \includegraphics[width=\columnwidth]{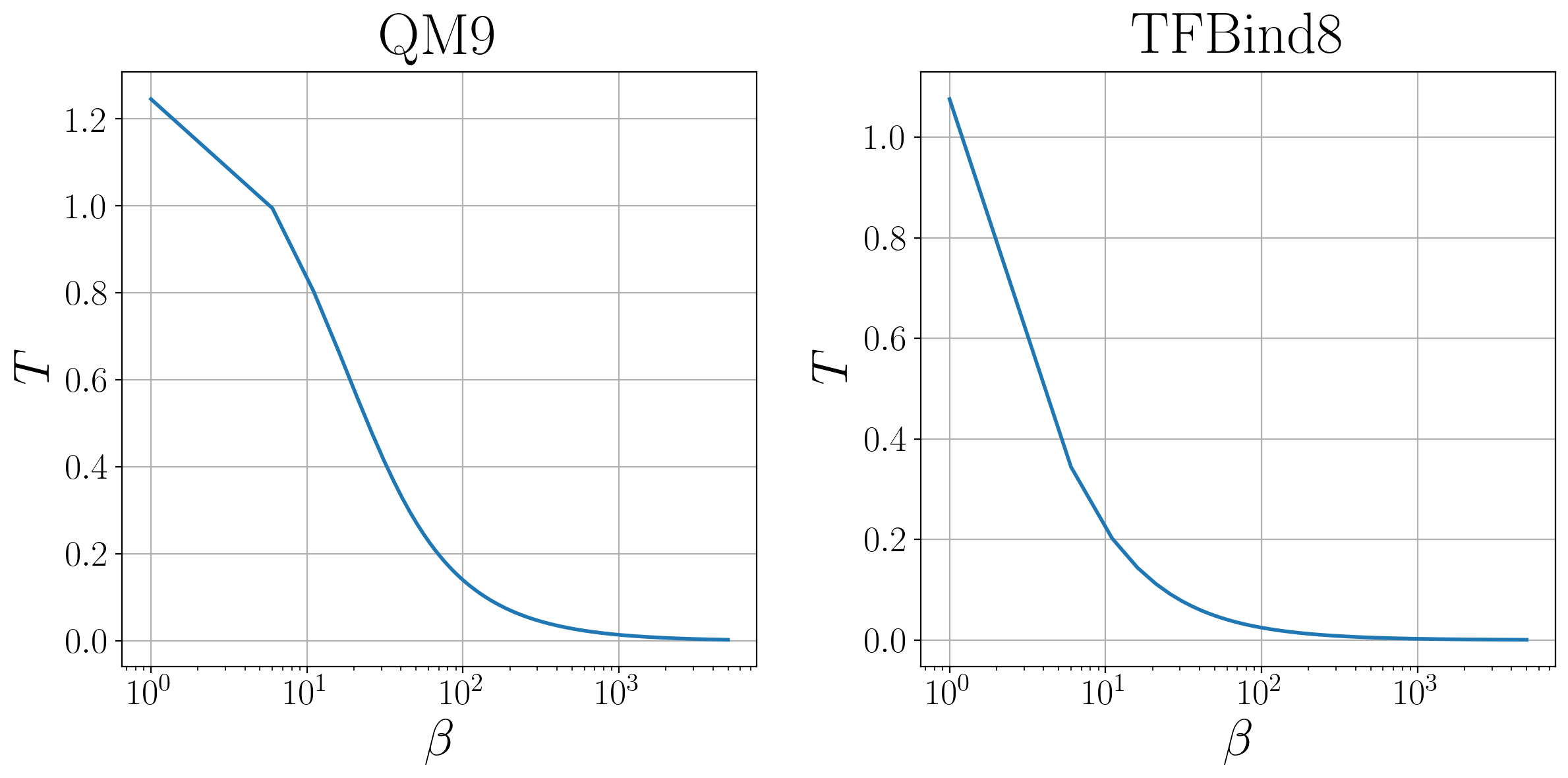}
    \vspace{-1em}
    \caption{Learned relationship between $\beta$ and $T$ captured by Logit-GFN. Logit-GFN suggests low softmax temperature when the policy is conditioned on high temperature (low $\beta)$.}
    \label{fig:beta_to_temp}
    \vspace{-1em}
\end{figure}

\textbf{Mapping from $\beta$ to $T$.} \Cref{fig:beta_to_temp} depicts the learned mapping from $\beta$ to $T$ in QM9 and TFBind8 tasks. We note that for both tasks, $T$ tends towards zero with the increase in $\beta$. It is a reasonable behavior since lower $T$ makes a narrow decision-making and accentuates the minor differences between probabilities.

\begin{figure*}[t]
\centering
\includegraphics[width=\textwidth]{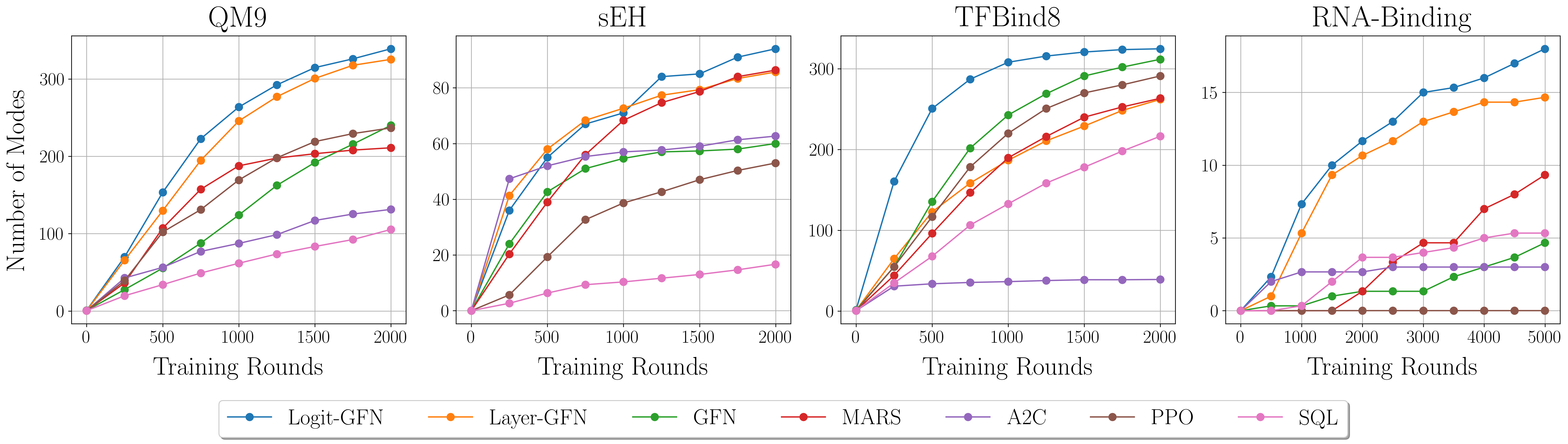}
\vspace{-15pt}
\caption{Number of modes discovered over training. Logit-GFN outperforms established baselines including Layer-GFN for all tasks.} 
\label{fig:modes}
\end{figure*}

\begin{figure*}[t]
\begin{minipage}[t]{\textwidth}
    \begin{subfigure}[t]{0.33\textwidth}
        \centering
        \includegraphics[width=\textwidth]{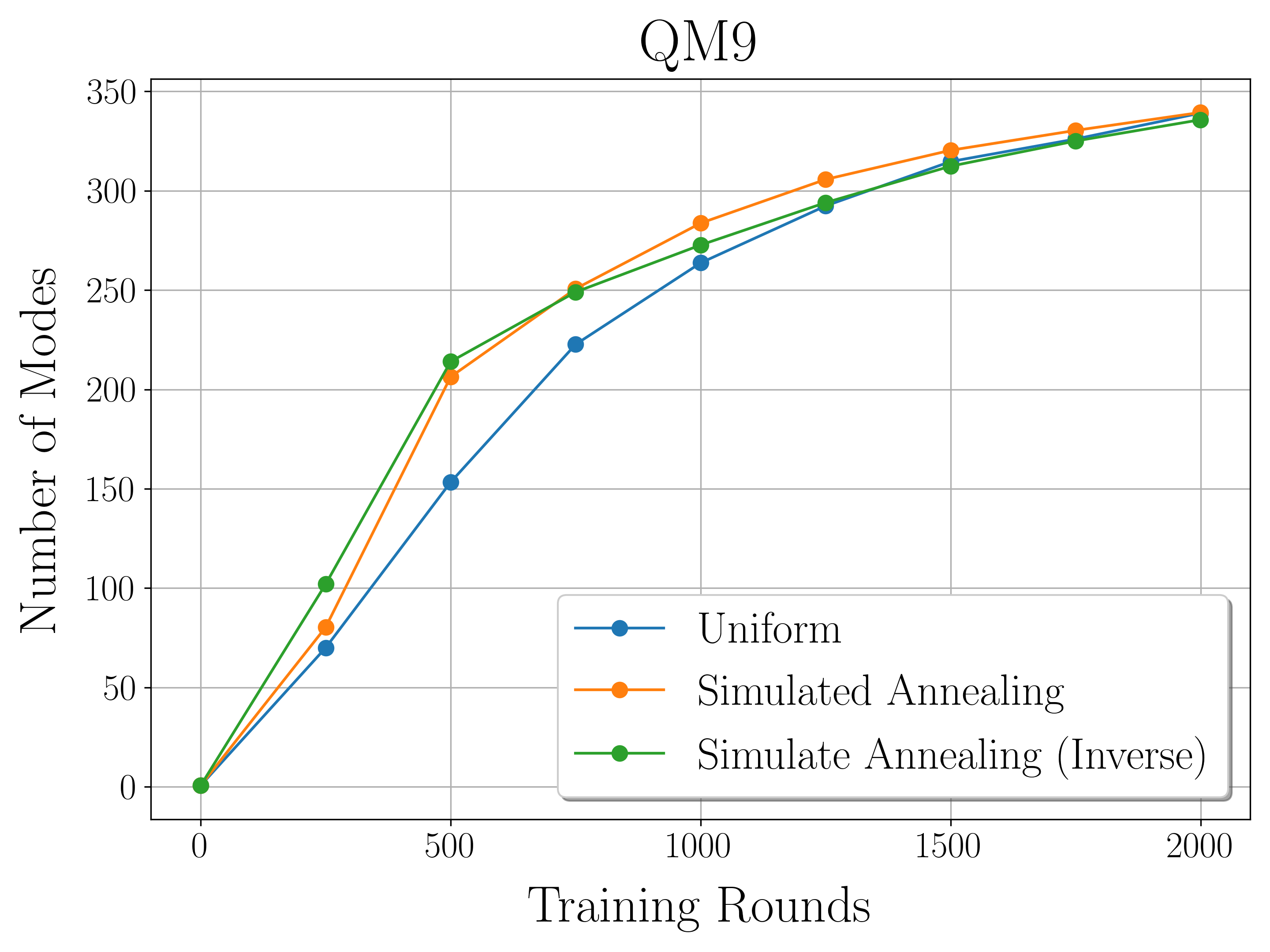}
        \label{fig:qm9_query_dist_ablation}
    \end{subfigure}
    \begin{subfigure}[t]{0.33\textwidth}
        \centering
        \includegraphics[width=\textwidth]{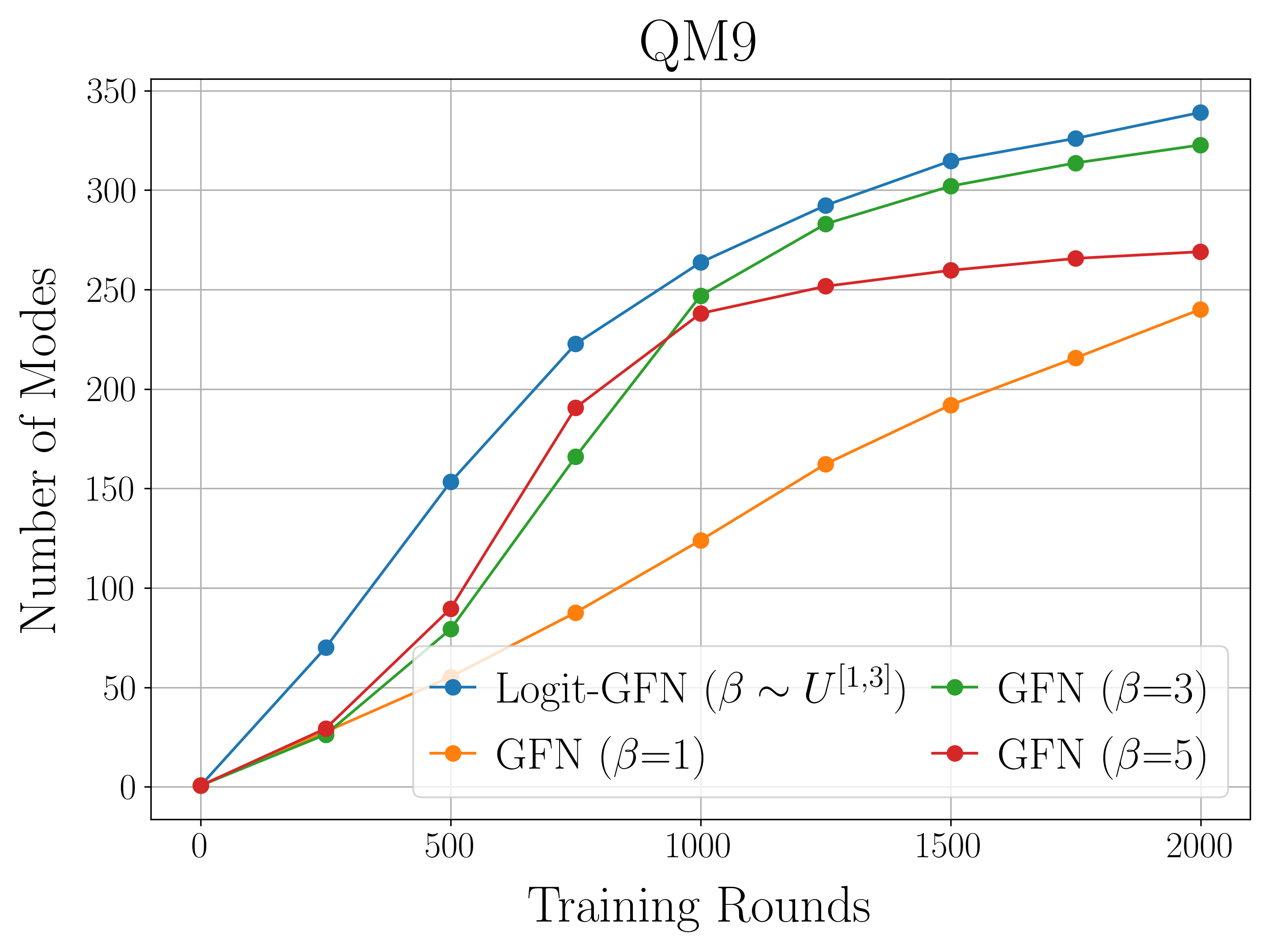}
        \label{fig:qm9_beta_ablation}
    \end{subfigure}
    \begin{subfigure}[t]{0.33\textwidth}
        \centering
        \includegraphics[width=\textwidth]{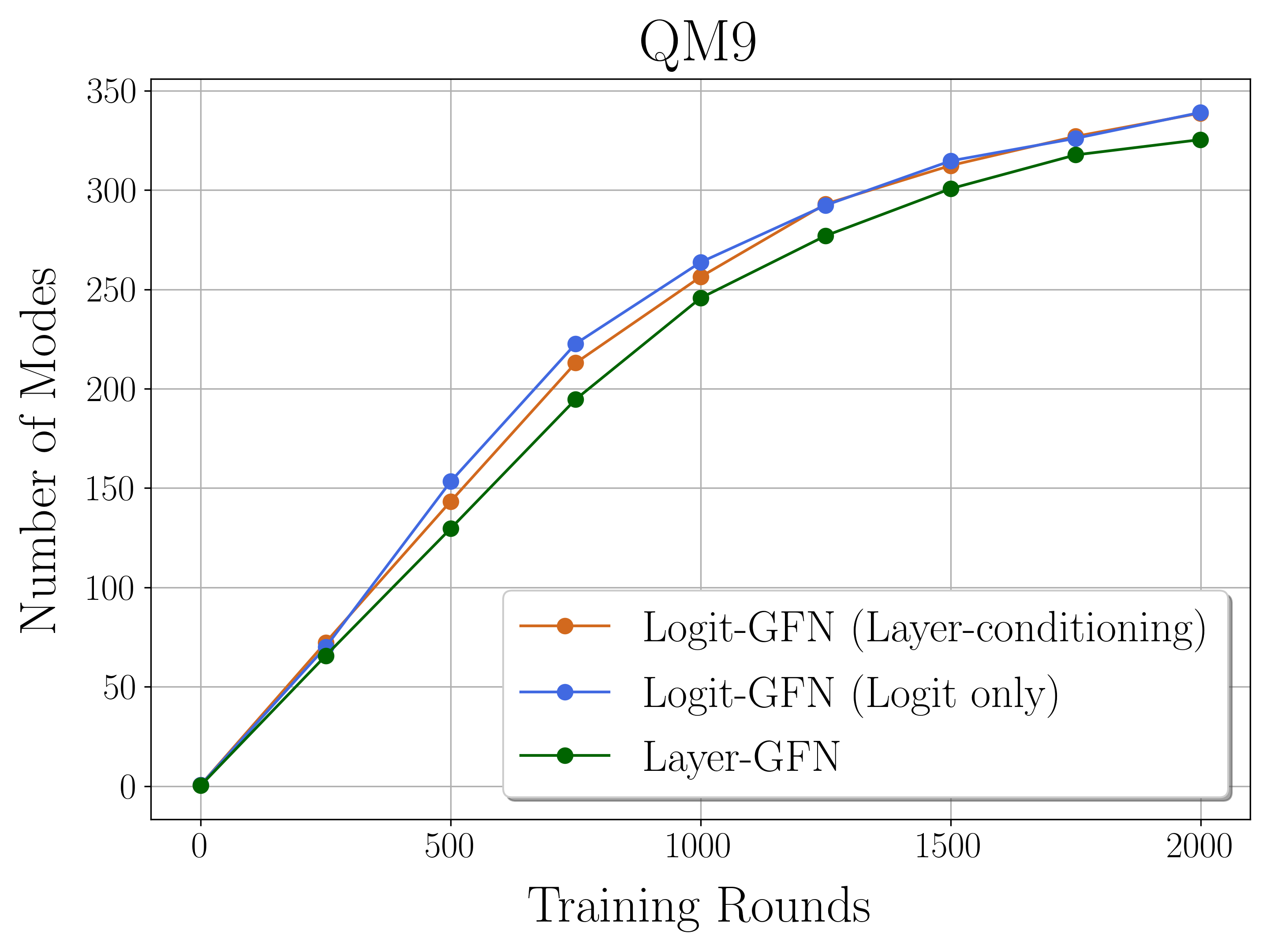}
        \label{fig:qm9_logit+layer_ablation}
    \end{subfigure}
    \vspace{-20pt}
    \caption{\textbf{Left}: Different distributions for sampling temperatures. The simulated annealing strategy is more effective than uniform sampling in online mode-seeking problems. \textbf{Middle}: Ablation on different $\beta$ for unconditional GFlowNets. Logit-GFN outperforms all GFlowNets trained on specific $\beta$. \textbf{Right}: Ablation on Layer-conditioning for Logit Scaling. Both outperforms Layer-GFN.}\label{ablation_studies_main}
\end{minipage}
\vspace{-1em}
\end{figure*}

\subsection{Evaluation of online mode seeking capability}
The preceding sections have demonstrated that Logit-GFN provides enhanced training stability and better controllability compared to unconditional GFlowNets and Layer-GFN. In this section, we aim to validate the usefulness of Logit-GFN in solving online mode-seeking problems. 

The online mode-seeking problem is an important problem in biochemical tasks as we require generating both diverse and high-reward samples due to the lack of robustness and misspecification of the estimated reward function \citep{bengio2021flow}. The use of temperature-conditional GFlowNet is particularly beneficial in this setting as we can more easily balance exploration and exploitation by querying different $\beta$ values, which is very useful for discovering a greater diversity of modes of the reward function.

\textbf{Baselines.} To evaluate the proposed methods, we first compare them with the established unconditional GFlowNet. We also consider MARS \cite{xie2020mars}, which is an MCMC-based method known to work well in this molecule sampling domain, and RL-based methods which include A2C with Entropy Regularization \cite{mnih2016asynchronous}, Soft Q-Learning \cite{haarnoja2018soft}, and PPO \cite{schulman2017proximal} as baselines. All experiments are conducted with three different random seeds to evaluate reliability.

\textbf{Distribution for sampling temperature.} Online mode seeking problems involve two stages of sampling temperatures, $P_{\text{train}}(\beta)$ and $P_{\text{exp}}(\beta)$. $P_{\text{train}}(\beta)$ is used for training temperature-conditional policy via off-policy manner, and $P_{\text{exp}}(\beta)$ is for sampling objects using trained policy in the online round. For both stages, we employ a uniform distribution as a default setting. We discuss various types of distribution, such as simulated annealing, in the next section.

\Cref{fig:modes} showcases the number of modes discovered during the training of our proposed approach compared to established baselines in four distinct biochemical tasks. Our method not only outperforms both the unconditional GFN and other reward maximization techniques in the number of modes uncovered but also exhibits remarkable efficiency in detecting these modes with a limited number of online evaluation rounds.
This underscores the effective capability of temperature-conditional GFlowNets to strike a balance between exploitation and exploration strategies.

\subsection{Different distribution for sampling temperatures} \label{beta_exploration}

We explore diverse distributions for sampling temperatures, especially during the exploration phase, namely $P_{\text{exp}}(\beta)$. One of the most widely used heuristics in combinatorial optimization is \textit{simulated annealing}, which gradually shifts the distribution of $\beta$ towards a high region \citep{kirkpatrick1983optimization}. Accordingly, we implement both simulated annealing and its inverse (shift distribution from high to low $\beta$ region). The effects of these exploration strategies in the QM9 task are illustrated in \Cref{ablation_studies_main} (Left). The simulated annealing strategy demonstrates high efficiency in discovering modes compared to the uniform strategy. 
While the inverse simulated annealing strategy exhibits fast convergence at the early stage of training rounds, but encounters a significant decrease in efficiency as the rounds advance.
It indicates that carefully adjusting the balance between exploration and exploitation can effectively enhance the performance of temperature-conditional GFlowNets for online mode-seeking problems. We conduct more extensive experiments on $P_{\text{exp}}(\beta)$ in \Cref{app:query_distribution}.

\subsection{Ablation studies}
We can adjust the $\beta$ of unconditional GFlowNets and train them independently to find the most effective configuration for a given task. \Cref{ablation_studies_main} (Middle) shows the number of modes discovered by GFlowNets trained with various $\beta$ values. We observe that all methods trained with fixed $\beta$ are outperformed by Logit-GFN, which trains conditional GFlowNets and balances exploitation and exploration with different $\beta$ values. 

We validate the effectiveness of layer-conditioning for Logit-GFN in online mode-seeking problems. As depicted in \Cref{ablation_studies_main} (Right), it seems layer-conditioning has a marginal effect on the performance of Logit-GFN, while still outperforming Layer-GFN. We discuss layer-conditioning in \Cref{app:ablation_logit+layer} in more detail.

We perform additional experiments on several design choices of temperature-conditional GFlowNets, such as thermometer embedding for Layer-GFN, the number of gradient steps per batch ($K$), and different GFlowNet training methods such as DB and SubTB in \Cref{app:thermo,app:ablation_K,app:ablation_training_obj}. 

\section{Conclusion}

We introduced the Logit-GFN, a novel architecture design that improves the training process of temperature-conditional GFlowNets. This approach addresses the numerical challenges of prior temperature-conditional GFlowNets by adopting the logit scaling net to scale the logits of policy directly. Empirical evaluations confirm that Logit-GFN can stabilize the training of temperature-conditional GFlowNets and exhibit strong generalization performance and effectiveness in diverse biochemical tasks. Additionally, while it demonstrates strong offline generalization and training stability in the tested biochemical tasks, further validation across a broader range of application domains is necessary to confirm its robustness and adaptability. A potential limitation of the Logit-GFN is that its effectiveness in very high-dimensional or highly complex generative tasks has not been fully explored. 

\section*{Impact Statement}

Our research is a part of the ongoing developments in the field of generative modeling and encompasses various societal implications typical of such advancements. However, we believe there are no specific consequences that require particular emphasis in this work.

\section*{Acknowledgement}

The authors acknowledge for the funding provided by CIFAR, NSERC, Samsung, FACS Acuité, and NRC AI4Discovery. This research was supported in part by computational resources from the Digital Research Alliance of Canada (\url{https://alliancecan.ca}), Mila (\url{https://mila.quebec}), and NVIDIA. J. Ko and W. Kim were supported by the National Research Foundation of Korea (NRF) grant funded by the Ministry of
Science and ICT (NRF-2022M3J6A1063021 and RS-2023-
00208980). This work was supported by Institute of Information \& communications Technology Planning \& Evaluation (IITP) grant funded by the Korea government(MSIT) (2022-0-01032, Development of Collective Collaboration Intelligence Framework for Internet of Autonomous Things)

\bibliography{main}
\bibliographystyle{icml2024}

\clearpage
\onecolumn
\appendix

\newpage

\section{Detailed experimental setting}
\label{app:exp-setting}

\subsection{Implementation}

In our GFlowNet implementations, we adhere closely to the methodologies outlined in \citep{shen2023towards}. We take the approach of re-implementing only those methods that do not already exist in the literature. 

All of our GFlowNet models incorporate the parametrization mapping of relative edge flow policy (SSR), as originally proposed by \cite{shen2023towards}.

When dealing with pairs of states $(s, s')$, we encode each state into a one-hot encoding vector and concatenate them as input for the forward and backward policy networks. For QM9 and sEH tasks, we employ a two-layer architecture with 1024 hidden units, while for the other tasks, we choose to use a two-layer architecture with 128 hidden units. Both forward and backward policies use the same architecture but with different parameters. We initialize $\log Z_{\theta}$ to 5.0 for the baseline implementation. 

For temperature-conditional GFlowNets, we introduce a two-layer MLP with a 32-dimensional hidden layer and a Leaky ReLU activation function for embedding inverse temperature $\beta$. For Layer-GFN, we use the output of the MLP as an embedding and concatenate with state embedding before passing forward policy networks. For Logit-GFN, we introduce an additional two-layer MLP, which has 32 hidden units and returns scale value, which represents $T$. To ensure that $T$ should be greater than zero, we use the Softplus activation function for the last layer. For parameterizing $Z_{\theta}(\beta)$, we use a two-layer MLP with a 32-dimensional hidden layer and Leaky ReLU.

\subsection{Hyperparameters}
\label{app:hyperparam}
Regarding the hyperparameters for GFlowNets, we also follow the initial settings proposed by \cite{shen2023towards} without alteration. Across all tasks, we employ the Adam optimizer \citep{kingma2014adam} with the following learning rates: $1\times10^{-2}$ for $Z_{\theta}$ and $1\times10^{-4}$ for both the forward and backward policy. Furthermore, we apply distinct reward exponents ($e$) and reward normalization constants for each task, following the guidelines suggested by \cite{shen2023towards}. As a result, we power the reward $e\times\beta$ times for temperature-conditional GFlowNets conditioned on $\beta$. \Cref{tab:hyperparam} summarizes the reward exponent and normalization constants for different task settings.

In the implementation of RL baselines, we utilize the same MLP architecture as employed in the GFlowNet baselines. The optimization of hyperparameters is achieved through a grid search approach on the QM9 task, with a focus on determining the optimal number of modes. For the A2C algorithm with entropy regularization, we segregate parameters for the actor and critic networks. The selected learning rate of $1\times10^{-4}$ is chosen from a range of options, including $\{1\times10^{-5}, 1\times10^{-4}, 1\times10^{-4}, 5\times10^{-3}, 1\times10^{-3}\}$, and we incorporate an entropy regularization coefficient of $1\times10^{-2}$ selected from $\{1\times10^{-4}, 1\times10^{-3}, 1\times10^{-2}\}$. In the case of Soft Q-Learning, we opt for a learning rate of $1\times10^{-4}$ selected from the same set of values: $\{1\times10^{-5}, 1\times10^{-4}, 1\times10^{-4}, 5\times10^{-3}, 1\times10^{-3}\}$. For the PPO algorithm, we introduce an entropy regularization term of $1\times10^{-2}$ and employ a learning rate of $1\times10^{-4}$, similarly chosen from $\{1\times10^{-5}, 1\times10^{-4}, 1\times10^{-4}, 5\times10^{-3}, 1\times10^{-3}\}$. The entropy regularization coefficient is selected from $\{1\times10^{-4}, 1\times10^{-3}, 1\times10^{-2}\}$.
\begin{table*}[h]
\centering
\caption{{GFlowNet hyperparameters for various tasks}}\label{tab:hyperparam}
% \resizebox{0.9\linewidth}{!}{
\begin{tabular}{lcccc}
\toprule
 Tasks & Reward Exponent  & Reward Normalization Constant\\
\midrule
QM9 & 5 & 100.0\\
sEH & 6 & 10.0\\
TFBind8 & 3  & 10.0\\
RNA-binding & 8  & 10.0\\
\bottomrule
\end{tabular}
% }
% \vspace{-10pt}
\end{table*}

\clearpage
\subsection{Experiment details}
\subsubsection{Training stability}\label{app:training_stability}
To evaluate the training stability, we compute the loss on samples generated from the forward policy conditioned on $\beta=5,000$. For each active round, we generate 32 samples for evaluating loss. After sampling, we train GFlowNets in an off-policy manner. We utilize the reward prioritized replay buffer (PRT) suggested in \citep{shen2023towards} to obtain samples for computing the loss of off-policy training for all models. We perform 1 gradient step per active round and use 32 samples from PRT to compute loss. During off-policy training, both Logit-GFN and Layer-GFN samples $\beta$ from the uniform distribution, i.e., $\beta\sim U^{[10, 50]}$.  

\subsubsection{Offline generalization}\label{app:offline_generalization}

For offline generalization, we train temperature-conditional GFlowNets in a relatively small $\beta$ during training and generate samples by querying with various $\beta$ from low to high values to see their generalization performance. For the QM9 task, we train temperature-conditional GFlowNets with $\beta\sim U^{[5, 10]}$ and query with $\beta=\{1, 5, 10, 50, 100, 500, 1000\}$. For TFBind8 and RNA-binding tasks, we train temperature-conditional GFlowNets with $\beta\sim U^{[10, 50]}$ and query with $\beta=\{1, 5, 10, 50, 100, 500, 1000, 5000\}$. As querying $\beta=5,000$ for the QM9 task leads to numerical instability due to the reward scaled to 100, we exclude the process from this experiment. Unconditional GFlowNets are trained with a fixed $\beta$ independently for each \(\beta\). All methods are run for 1,000 training steps and we generate 2,048 samples from the trained policy to measure the performance.

\subsubsection{Online mode seeking}
For online mode-seeking problems, we run experiments with $T=2,000$ training rounds for QM9, sEH, and TFBind8 and $T=5,000$ training rounds for RNA-Binding tasks. As sEH and RNA-binding tasks have a large combinatorial space to explore, we apply PRT for all GFlowNets methods to enhance sample efficiency. For each training round, we collect 32 online samples. During the off-policy training phase, we sample 32 samples from the replay buffer and update the policy via gradient descent $K$ times, where $K$ holds a significant role in the training of GFlowNets. A low number of gradient updates before sampling new trajectories can result in underfitting, while an excessive number of gradient updates may lead to overfitting. In contrast to the unconditional GFlowNets, we observed that temperature-conditional GFlowNets require more training iterations at each round to adapt to various temperature conditions effectively. We present ablations on $K$ in \Cref{app:ablation_K} with more details.

For online mode-seeking problems, the distribution of $\beta$ for training and exploration is crucial. As a default setting, we choose a uniform distribution for both the training and exploration phases. We summarize the distribution we used for different biochemical tasks in \Cref{tab:query_dist}.
\begin{table*}[h]
\centering
\caption{{Temperature Distributions of Temperature-conditioned GFlowNets for various tasks}}\label{tab:query_dist}
% \resizebox{0.9\linewidth}{!}{
\begin{tabular}{lcccc}
\toprule
 Tasks & Training Phase, $P_{\text{train}(\beta)}$ & Exploration Phase, $P_{\text{exp}(\beta)}$ & $K$\\
\midrule
QM9 & $U^{[1, 3]}$ & $U^{[1, 3]}$ & 4\\
sEH & $U^{[1, 5]}$ & $U^{[1, 5]}$ & 5\\
TFBind8 & $U^{[1, 3]}$ & $U^{[1, 3]}$ & 4\\
RNA-Binding & $U^{[2, 3]}$ & $U^{[2, 3]}$ & 1\\
\bottomrule
\end{tabular}
% }
% \vspace{-10pt}
\end{table*}

\clearpage
\subsection{Offline dataset details}\label{app:offline_dataset}

We verify the controllability of Logit-GFN with offline generalization. We prepare a sub-optimal offline dataset similar to offline model-based optimization for training GFlowNets. We describe the details of the dataset for each task.

\begin{itemize}
    \item \textbf{QM9}: In QM9 task, we build an offline dataset $\mathcal{D}$ using under 50th percentile data, which consists of 29,382 samples.
    \item \textbf{TFBind8}: In TFBind8 task, we follow the method suggested in Design-bench \citep{trabucco2022design}. We build an offline dataset $\mathcal{D}$ using under 50th percentile data, which consists of 32,898 samples.
    \item \textbf{RNA-Binding}: In the RNA-binding task, we follow the method suggested in BootGen \citep{kim2023bootstrapped}. We prepare an offline dataset consisting of 5,000 randomly generated RNA sequences. 
\end{itemize}

\subsection{Mode metrics details}\label{app:mode_metrics}
We define the measure of the number of modes in a sampled dataset as the count of data points with a reward above a specified threshold, where each data point is dissimilar from the others. The threshold is determined by the reward of the top 0.5\% of samples Additionally, we use the Tanimoto diversity metric for molecule optimization tasks to measure dissimilarity. We only accept samples as modes that are far away from previously accepted samples in terms of Tanimoto diversity. We set the threshold as 0.5. For RNA-binding tasks, we define mode as a local optimum among its 2-hamming ball neighborhoods. For the TFBind8 task, since there is already a well-pre-defined set of modes considering both optimality and diversity, we use that pre-defined set for evaluation.

\clearpage
\subsection{Additional results in offline generalization}
\begin{figure*}[h!]
    \centering
    \includegraphics[width=0.95\textwidth]{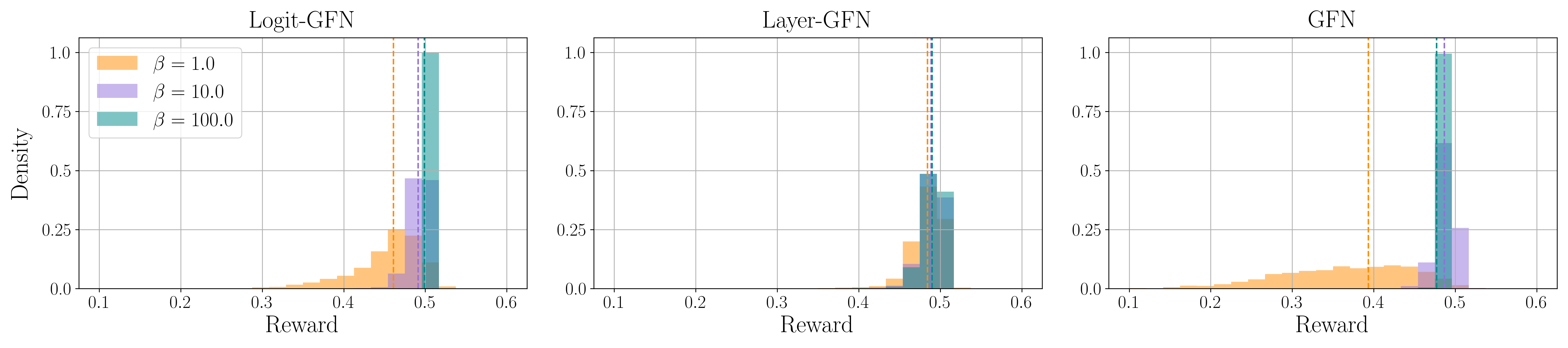}
    \vspace{-10pt}
    \caption{Reward distribution of samples generated from between Temperature-conditional GFlowNets and unconditional GFlowNet in offline generalization. Task is QM9.}
    \label{fig:qm9str_histogram}
\end{figure*}

\begin{figure*}[h!]
    \centering
    \includegraphics[width=0.95\textwidth]{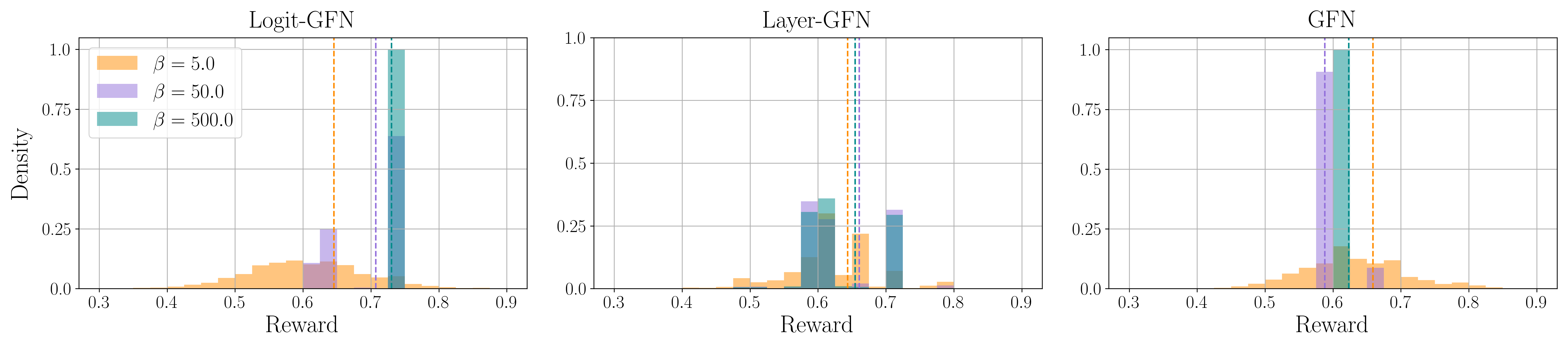}
    \vspace{-10pt}
    \caption{Reward distribution of samples generated from between Temperature-conditional GFlowNets and unconditional GFlowNet in offline generalization. Task is RNA-binding.}
    \label{fig:rna_histogram}
\end{figure*}

\begin{figure*}[h!]
    \centering
    \includegraphics[width=0.95\textwidth]{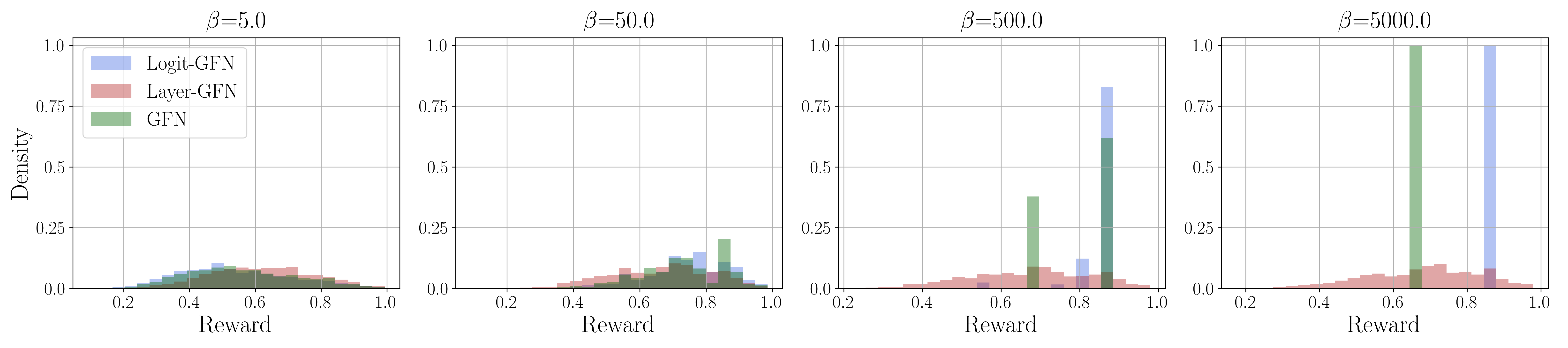}
    \vspace{-10pt}
    \caption{Reward distribution of samples generated from between temperature-conditional GFlowNets and unconditional GFlowNet in offline generalization. Task is TFBind8.}
    \label{fig:tfbind8_histogram_transpose}
\end{figure*}

We also visualize the reward distribution obtained from QM9 and RNA-binding tasks in \Cref{fig:qm9str_histogram,fig:rna_histogram}, which show the reward distribution of samples when querying with different $\beta$ values. For all tasks, it is observed that Logit-GFN focuses on high-scoring samples when querying with high $\beta$, and reacts more adaptively to changes in $\beta$ compared to Layer-GFN. The performance of Unconditional GFN methods degrades when trained with excessively high $\beta$ values, highlighting the need for temperature-conditional GFlowNets with enhanced generalization capabilities. Additionally, histograms are presented in \Cref{fig:tfbind8_histogram_transpose} to facilitate a direct comparison between different GFlowNets using the same $\beta$ values.

Note that our objective is not to achieve state-of-the-art performance in offline MBO setting \citep{trabucco2022design}. Our aim is to verify that Logit-GFN can generate out-of-distribution high-scoring samples by querying with high $\beta$ through generalization capability learned from training with various $\beta$ values.

\clearpage
\section{Additional Experiments}
\subsection{Study on thermometer encoding}\label{app:thermo}

It is often challenging to retrieve useful features from scalar input, which in our case is \(\beta\). One way to overcome this is to utilize an additional encoding function in the very first of the embedding function. Among different encoding techniques, \citet{buckman2018thermometer} introduced the thermometer encoding for such inputs to increase the robustness of neural networks, and \citet{jain2023multi} proved its effectiveness in preference-conditional GFlowNets. 

\begin{figure*}[h]
    \centering
    \includegraphics[width=0.95\textwidth]{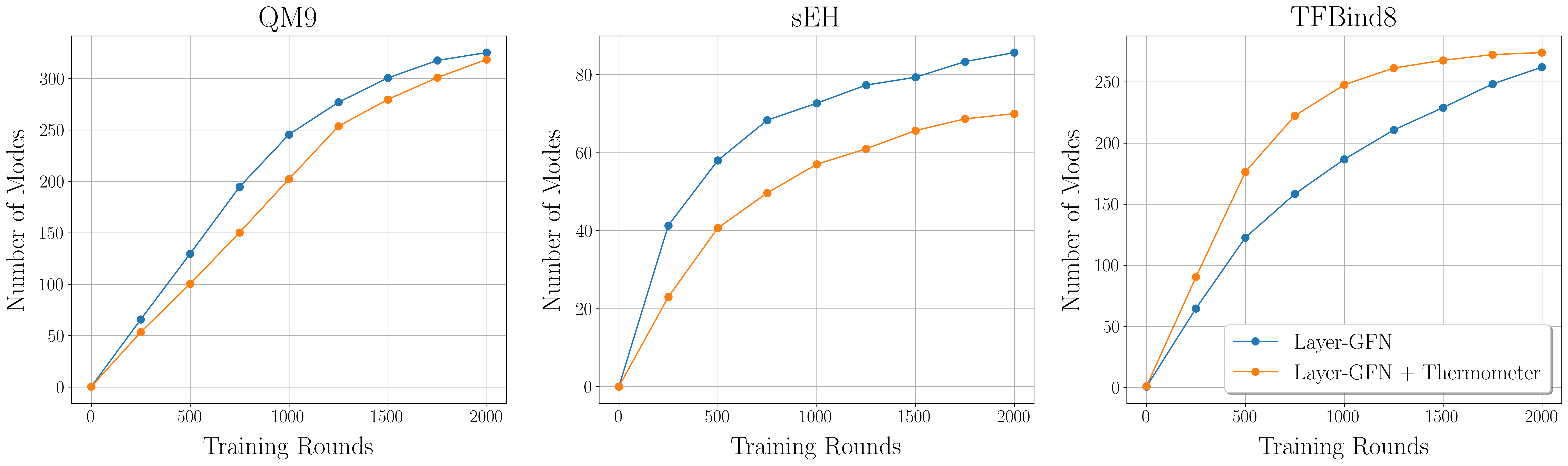}
    \caption{Ablation on Thermometer Encoding for Layer-GFN (Number of Modes)}\label{fig:thermometer_ablation}
\end{figure*}

Therefore, we ablate the effectiveness of thermometer encoding for Layer-GFN in online mode-seeking problems. \Cref{fig:thermometer_ablation} shows the performance of Layer-GFN with and without thermometer encoding in various tasks. As shown in the figure, Layer-GFN with thermometer encoding outperforms naive Layer-GFN in TFBind8 but underperforms in QM9 and sEG. It seems applying thermometer encoding is not always useful, and its effectiveness varies across tasks. Therefore, we do not use thermometer encoding for Layer-GFN in the main experiments.

\begin{figure*}[h]
    \centering
    \includegraphics[width=0.95\textwidth]{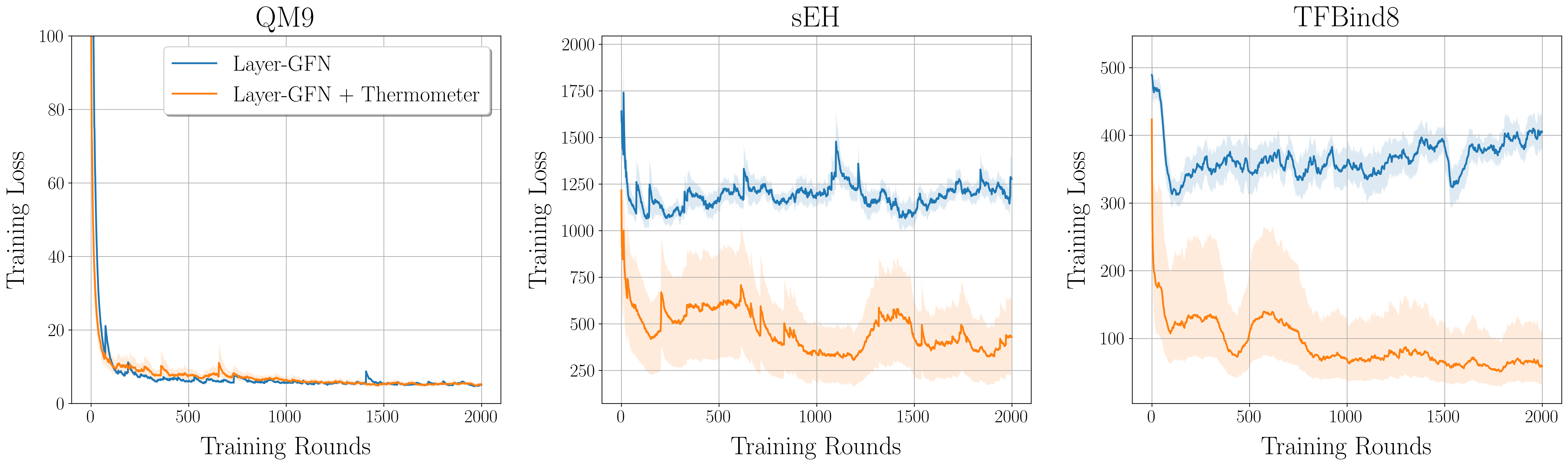}
    \caption{Ablation on Thermometer Encoding for Layer-GFN (Loss Curve)}\label{fig:thermometer_ablation_loss}
\end{figure*}

We also analyze the loss curve across training Layer-GFN with and without thermometer encoding. As \Cref{fig:thermometer_ablation_loss} illustrates, thermometer encoding generally stabilizes the training of Layer-GFN while it is not always helpful in seeking modes.

\clearpage
\subsection{Ablation on $\beta$ of unconditional GFlowNets}\label{app:ablation_beta}
\begin{figure}[h]
\begin{minipage}[t]{\textwidth}
    \begin{subfigure}[t]{0.5\textwidth}
        \centering
        \includegraphics[width=\textwidth]{figures/qm9str_modes_div_threshold_05_beta_ablation.png}
        \label{fig:qm9_beta_ablation_copy}
    \end{subfigure}
    \begin{subfigure}[t]{0.5\textwidth}
        \centering
        \includegraphics[width=\textwidth]
        {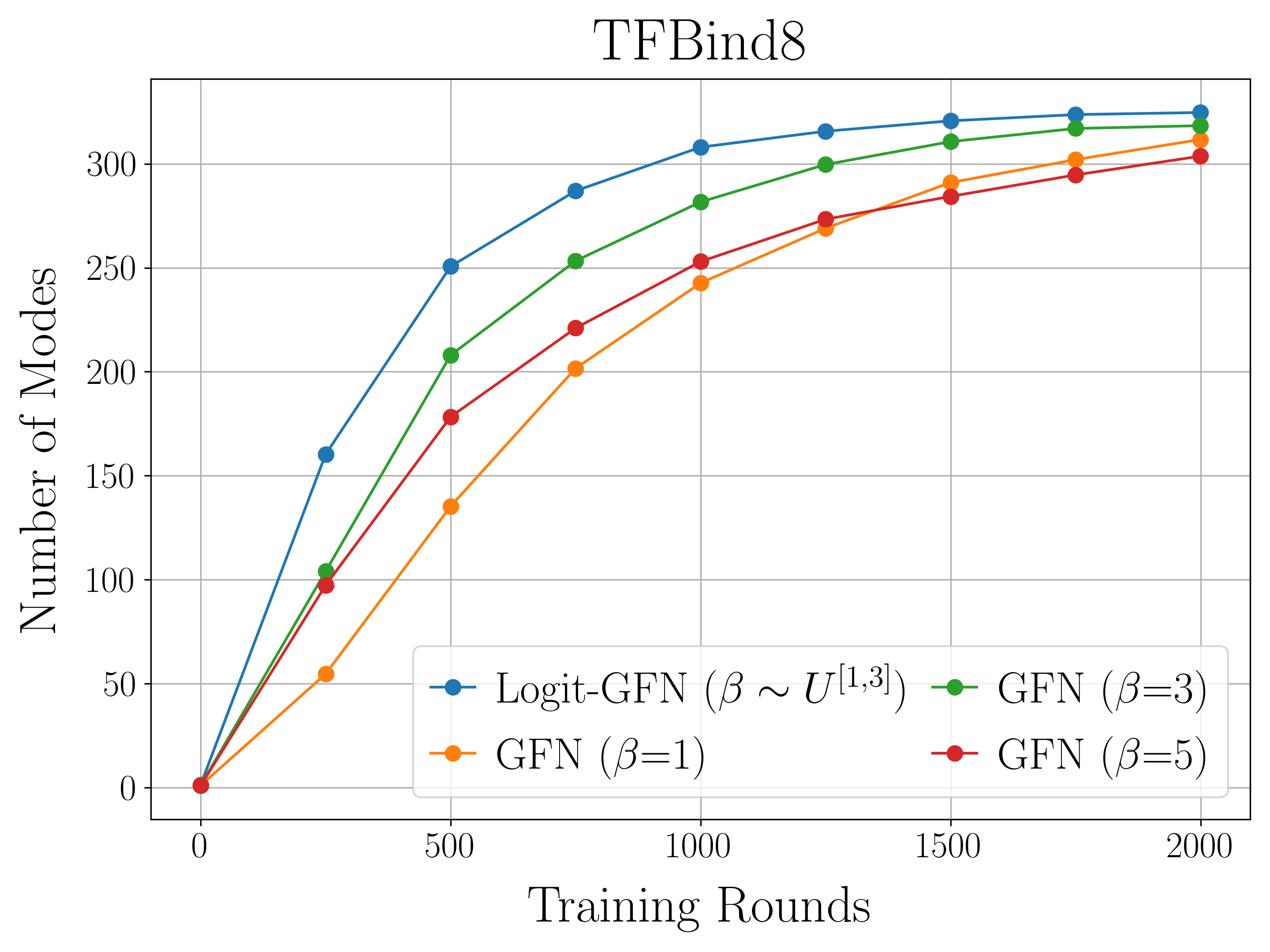}
        \label{fig:tfbind8_beta_ablation}
    \end{subfigure}
    \vspace{-20pt}
\end{minipage}
\caption{Ablation on different $\beta$ for unconditional GFlowNets}\label{fig:beta_ablation}
\end{figure}

In the main section, we ablate on $\beta$ of unconditional GFlowNets in QM9. We also conduct similar experiments in other biochemical tasks. \Cref{fig:beta_ablation} shows the performance of unconditional GFlowNets with different $\beta$ in QM9 and TFBind8 tasks. As shown in the figure, Logit-GFN outperforms all the other unconditional GFlowNets tailored by a fixed $\beta$ value, demonstrating the effectiveness of Logit-GFN in seeking online modes.

\subsection{Ablation on number of steps per batch ($K$)}\label{app:ablation_K}
\begin{figure}[h]
\begin{minipage}[t]{\textwidth}
    \begin{subfigure}[t]{0.5\textwidth}
        \centering
        \includegraphics[width=\textwidth]{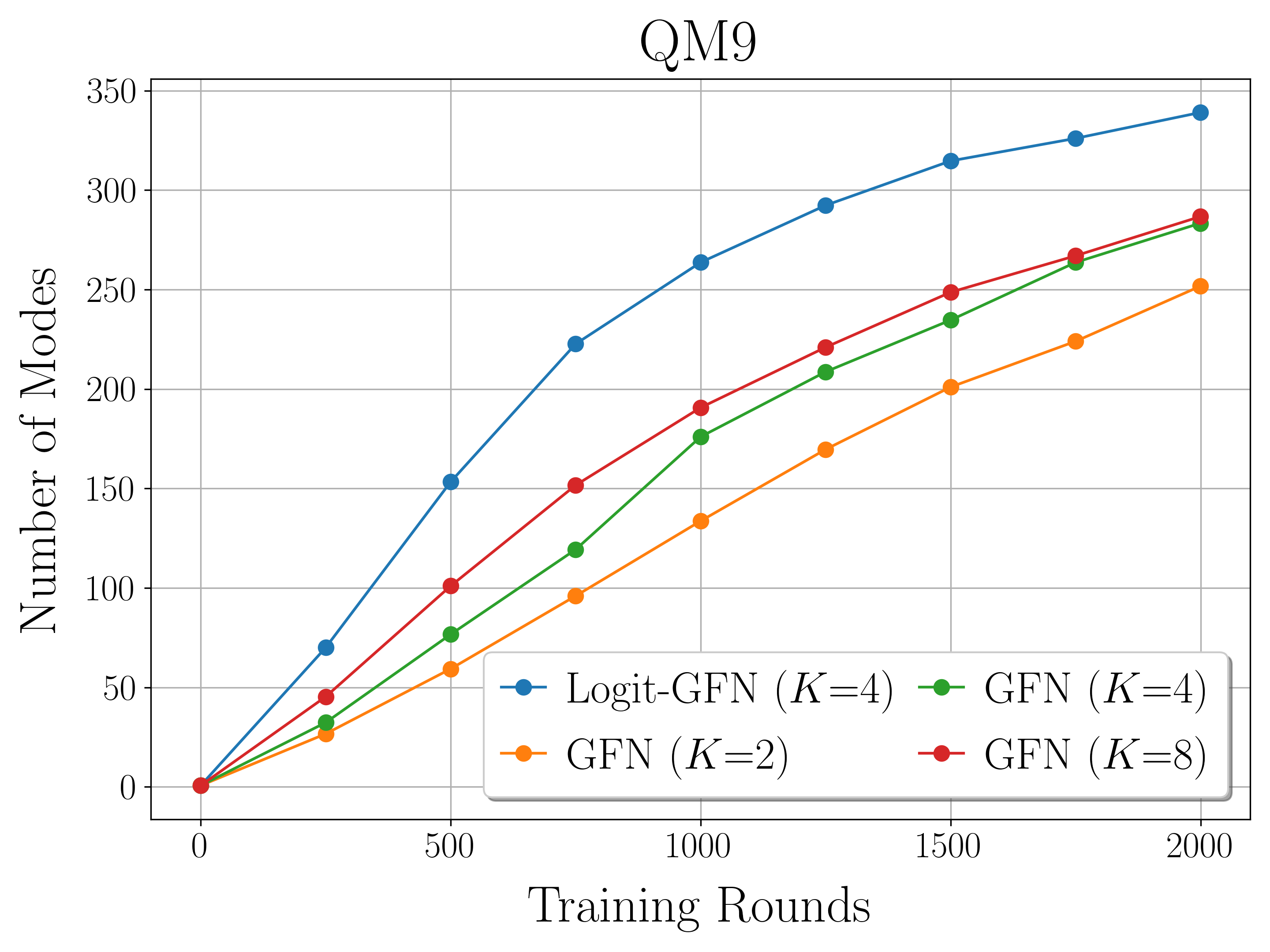}
        \label{fig:qm9_inner_loop_ablation}
    \end{subfigure}
    \begin{subfigure}[t]{0.5\textwidth}
        \centering
        \includegraphics[width=\textwidth]
        {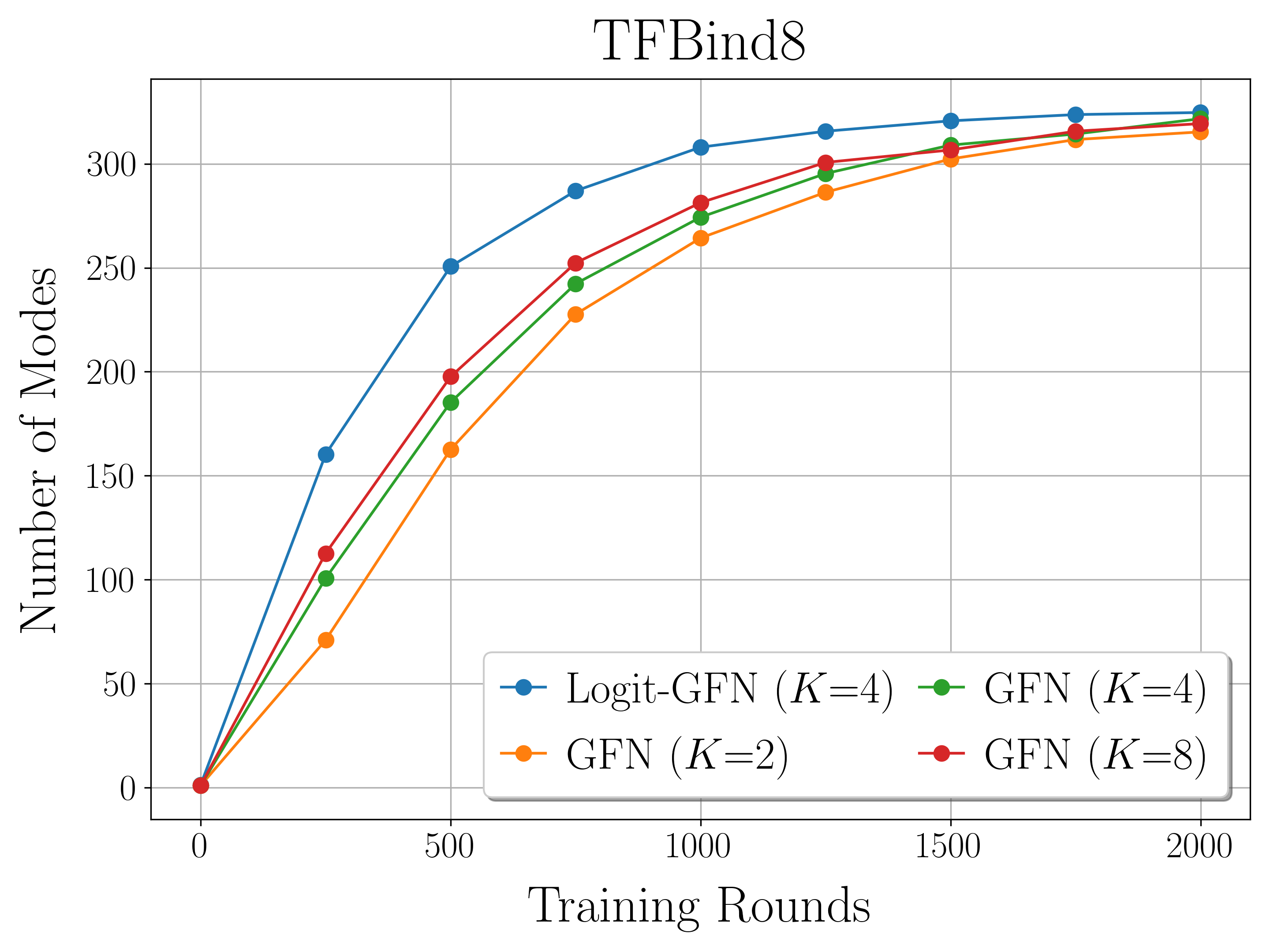}
        \label{fig:tfbind8_inner_loop_ablation}
    \end{subfigure}
    \vspace{-20pt}
\end{minipage}
\caption{Ablation on different $K$ for unconditional GFlowNets}\label{fig:inner_loop_ablation}
\end{figure}

The number of gradient steps per batch, $K$, holds a significant role in training GFlowNets. As a default setting, we choose $K=1$ for unconditional GFlowNets as increasing $K$ can lead to overfitting on current observations and converge to the local optimum. \Cref{fig:inner_loop_ablation} shows the performance of unconditional GFlowNets with increasing $K$. As illustrated in the figure, even if we increase $K$, Logit-GFN still outperforms unconditional GFlowNets by a large margin.

\clearpage
\subsection{Ablation on different GFlowNet training methods}\label{app:ablation_training_obj}
Apart from TB, there are other loss functions for training GFlowNets that satisfy \Cref{eq:gfn}, such as detailed balance (DB) and sub-trajectory balance (SubTB). In principle, our logit-scaling trick can accommodate any loss function. To verify this, we evaluate the effectiveness of our method with DB and SubTB training objectives as well. For SubTB, we use $\lambda=0.9$. To implement DB and SubTB objectives, we need to predict state flow. We introduce a temperature-conditional state flow $F(s;\beta,\theta):\mathcal{S}\times\mathbb{R}\rightarrow\mathbb{R}$ to parameterize state flow for Logit-GFN. We choose a two-layer MLP with a 32-dimensional hidden layer and a Leaky ReLU activation. The temperature-conditioned version of DB and SubTB can be written as:
\begin{align}
\label{eq:loss_variant}
    &\mathcal{L}_{\text{DB}}(\theta;\mathcal{D}) 
    = \mathbb{E}_{P_{\text{train}}(\beta)}\mathbb{E}_{P_{\mathcal{D}}(\tau)}\left[\sum_{t=1}^{n-1}\left(\log\frac{F(s_{t-1};\beta, \theta)P_F(s_{t}\vert s_{t-1};\beta,\theta)}{F(s_{t};\beta, \theta)P_B(s_{t-1}\vert s_{t};\beta,\theta)}\right)^2 + \left(\log\frac{F(s_{n-1};\beta, \theta)P_F(x\vert s_{n-1};\beta,\theta)}{R(x)^{\beta}P_B(s_{n-1}\vert x;\beta,\theta)}\right)^2\right].\\
    &\mathcal{L}_{\text{SubTB}}(\theta;\mathcal{D}) 
    = \mathbb{E}_{P_{\text{train}}(\beta)}\mathbb{E}_{P_{\mathcal{D}}(\tau)}\left[\frac{\sum_{0\leq i<j\leq n}\lambda^{j-i}\mathcal{L}_{\text{SubTB}}^{i, j}(\theta;\tau)}{\sum_{0\leq i<j\leq n}\lambda^{j-i}}\right]. \\ \nonumber \\
    &\mathcal{L}_{\text{SubTB}}^{i, j}(\theta;\tau) 
    =
    \begin{cases*}
        \left(\log\frac{F(s_{i};\beta, \theta)\prod_{t=i+1}^{n}P_F(s_{t}\vert s_{t-1};\beta,\theta)}{R(x)^{\beta}\prod_{t=i+1}^{n}P_B(s_{t-1}\vert s_{t};\beta,\theta)}\right)^2 & if $j=n$,\\
        \left(\log\frac{F(s_{i};\beta, \theta)\prod_{t=i+1}^{j}P_F(s_{t}\vert s_{t-1};\beta,\theta)}{F(s_{j};\beta, \theta)\prod_{t=i+1}^{j}P_B(s_{t-1}\vert s_{t};\beta,\theta)}\right)^2 & otherwise. 
    \end{cases*}
\end{align}

\begin{figure}[h]
\centering
\begin{minipage}[t]{0.8\textwidth}
    \begin{subfigure}[t]{0.5\textwidth}
        \centering
        \includegraphics[width=\textwidth]{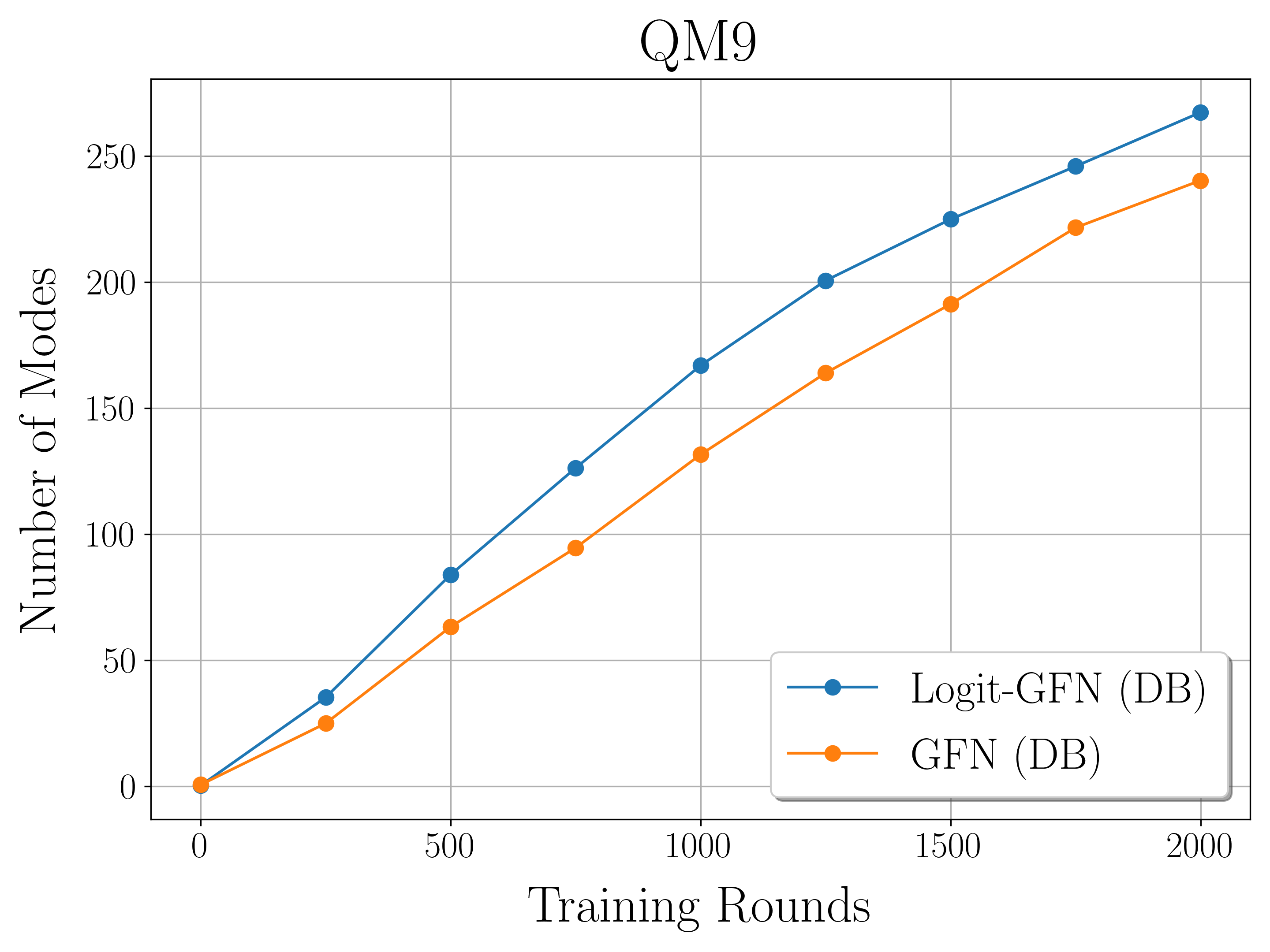}     \label{fig:qm9_loss_ablation_db}
    \end{subfigure}
    \begin{subfigure}[t]{0.5\textwidth}
        \centering
        \includegraphics[width=\textwidth]{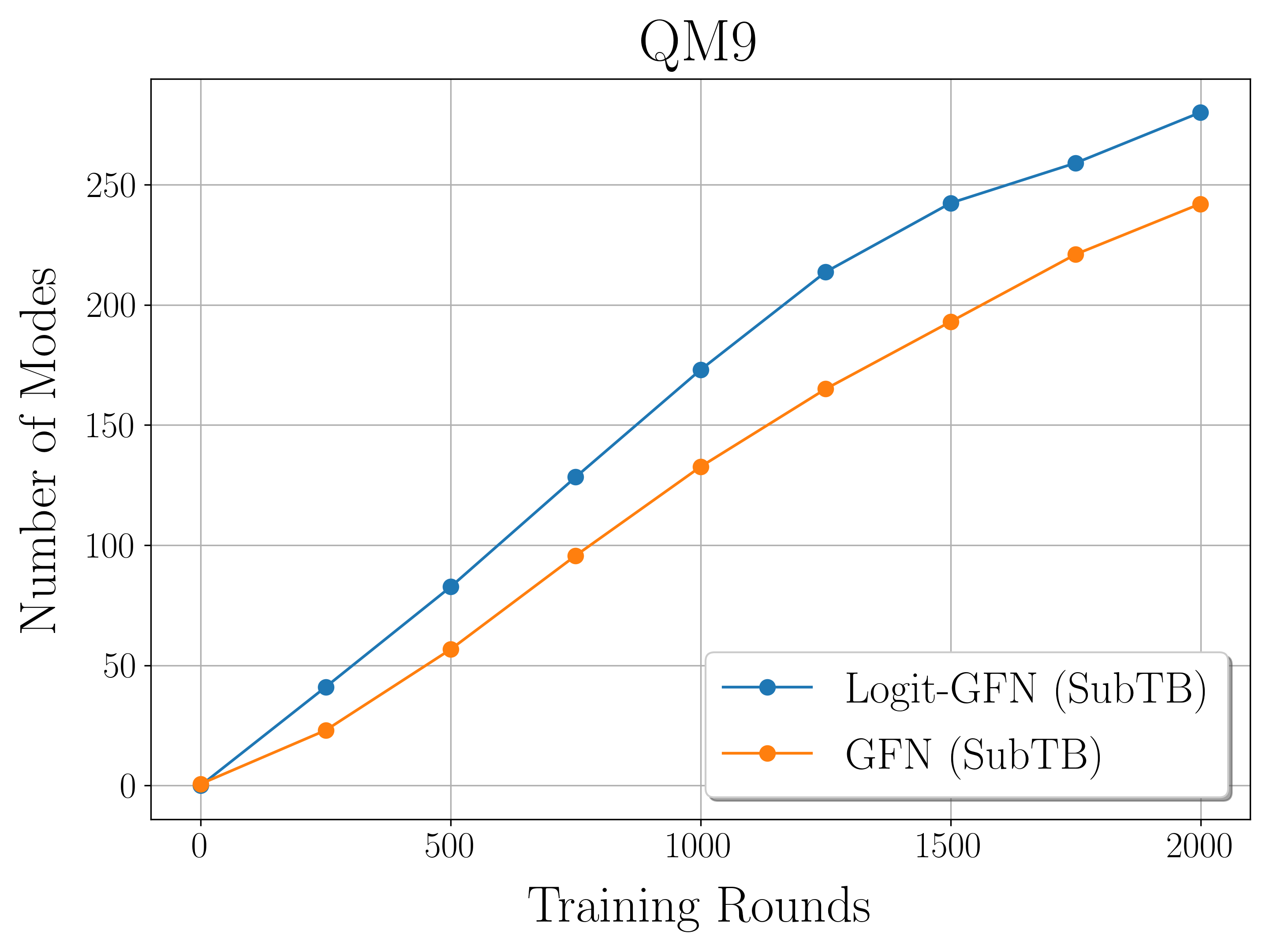}
\label{fig:qm9_loss_ablation_subtb}
    \end{subfigure}
\end{minipage}
\vspace{-20pt}
\caption{Ablation on different GFlowNet Training Methods in QM9 Task.}\label{fig:loss_ablation}
\end{figure}

\begin{figure}[h]
\centering
\begin{minipage}[t]{0.8\textwidth}
    \begin{subfigure}[t]{0.5\textwidth}
        \centering
        \includegraphics[width=\textwidth]{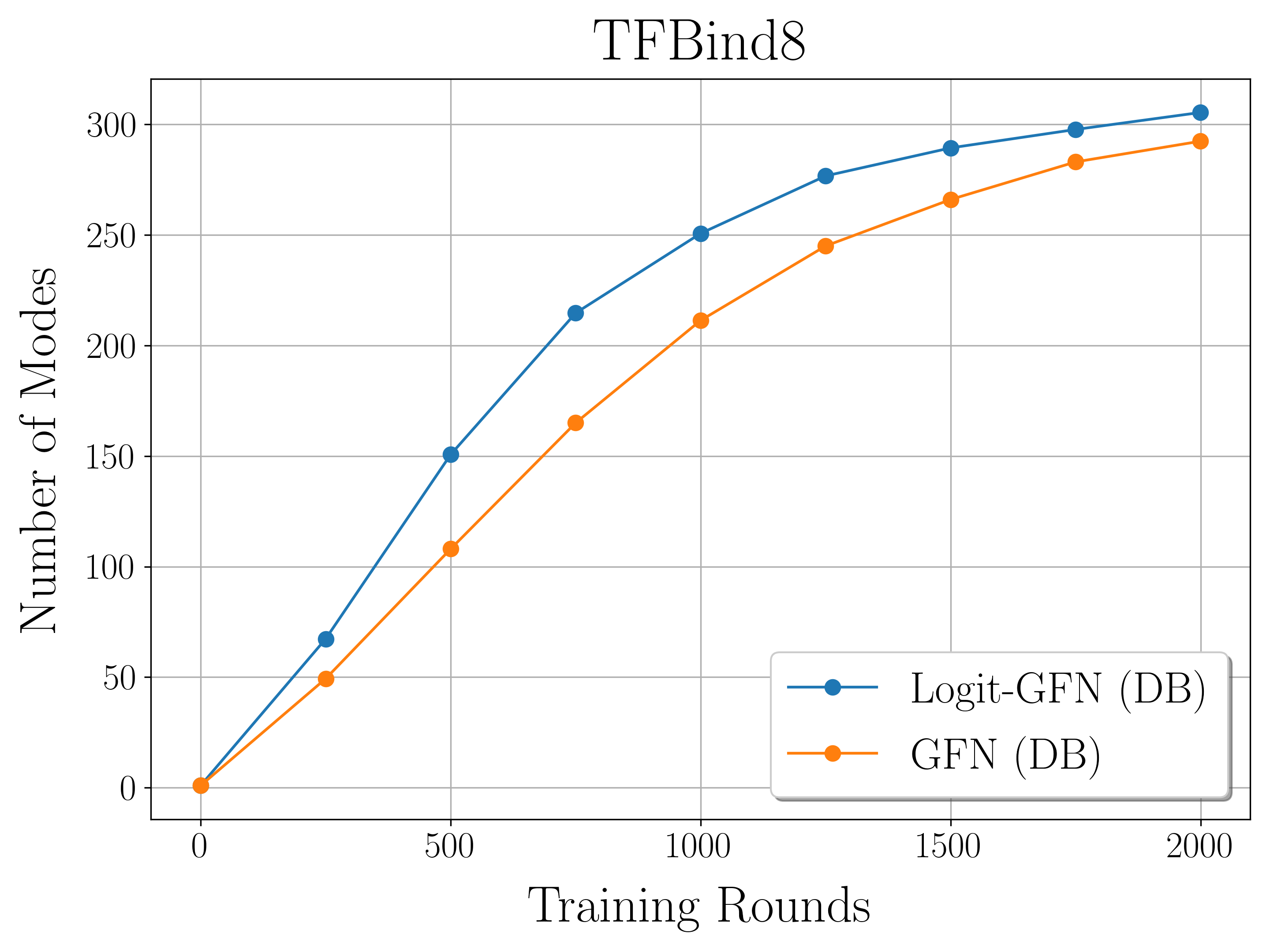}
        \label{fig:tfbind8_loss_ablation_db}
    \end{subfigure}
    \begin{subfigure}[t]{0.5\textwidth}
        \centering
        \includegraphics[width=\textwidth]{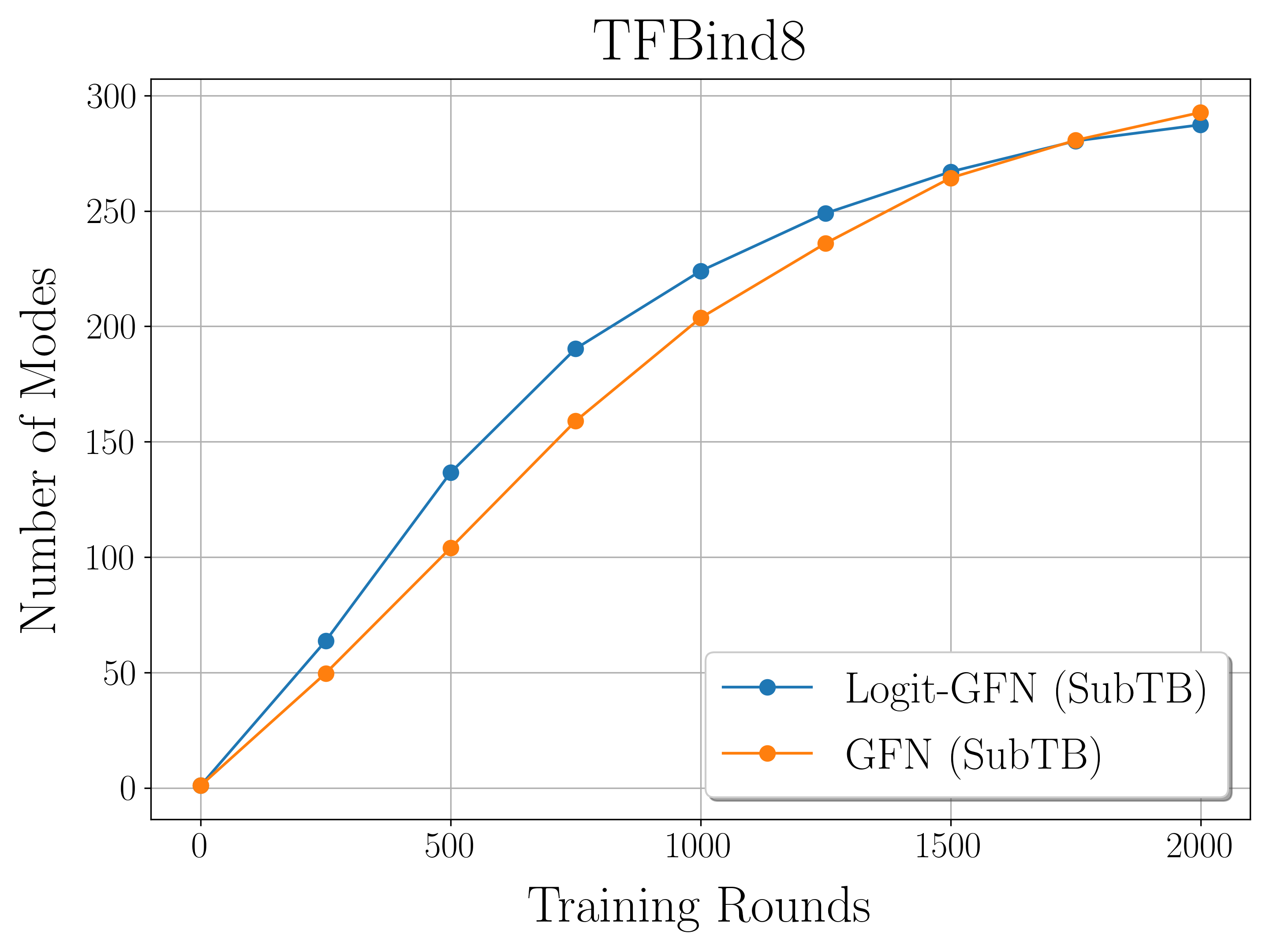}
        \label{fig:tfbind8_loss_ablation_subtb}
    \end{subfigure}
\end{minipage}
\vspace{-20pt}
\caption{Ablation on different GFlowNet Training Methods in TFBind8 task}\label{fig:loss_ablation_tfbind8}
\end{figure}

\Cref{fig:loss_ablation,fig:loss_ablation_tfbind8} shows the number of discovered modes over training with different objectives in QM9 and TFBind8. Logit-GFN consistently outperforms unconditional GFN across various objectives, demonstrating its effectiveness.

\clearpage

\subsection{Experiments on non-biochemical tasks}
As GFlowNets are not limited to biochemical domains, though they show significant promise in that area, our approach, Logit-GFN, is capable of performing well in other non-biochemical domains. One of the most critical areas for application is combinatorial optimization (CO). Specifically, we trained Logit-GFN in the graph combinatorial optimization domain, including tasks such as maximum independent sets (MIS) and maximum clique (MC) on top of detailed balance (DB) GFlowNet. We followed the implementation provided in \citet{zhang2023let} \footnote{\url{https://github.com/zdhNarsil/GFlowNet-CombOpt}}. As shown in \Cref{table:table_co}, Logit-GFN outperforms GFlowNet across various experiments within the combinatorial optimization domain, demonstrating its generalizability.

% \begin{figure*}[!h]
% \centering
\begin{table}[h!]
  \centering
    \caption{Experiments in non-biochemical domains. We sample 20 solutions for each graph configuration and take the best result.}
    \resizebox{0.6\linewidth}{!}{
    \begin{tabular}{lllll}
    \toprule
     &MIS (Small)&MIS (Large)&MC (Small)&MC (Large)\\
    \midrule
    Logit-GFN (DB) & \textbf{19.20 ± 0.07} & \textbf{35.65 ± 0.61} & \textbf{16.54 ± 0.20} & \textbf{34.08 ± 3.15}\\
    GFN (DB) & 17.93 ± 0.69 & 34.19 ± 0.37 & 15.56 ± 0.11 & 29.36 ± 0.01\\
    \bottomrule
    \end{tabular}}
  \label{table:table_co}
\end{table}
\clearpage

\section{Logit-GFN with Layer-conditioning}\label{app:ablation_logit+layer}
We investigate the impact of incorporating layer-conditioning into Logit-GFN. In our approach, the embedding produced by the encoder of the logit scaling network, denoted as $f^1_{\theta}$, is concatenated with the state embedding. This combined embedding is then input into the forward policy networks. The process of concatenation is illustrated in \Cref{fig:main}.

\subsection{Experiments on Gridworld}
We conduct experiments on Gridworld to quantify the discrepancy between target and sampled distributions. In these experiments, we measure the L1 distance between the true distributions and the generative distributions, following the methodology outlined in previous works \cite{bengio2021flow}.

As shown in the tables below, Logit-GFN (both Layer-conditioning and Logit-only) achieves more accurate sampling compared to Layer-GFN and GFlowNet. Moreover, layer-conditioning yields the smallest L1 distance from the true underlying distribution. This suggests that integrating logit-scaling with the layer-conditioning method enhances the expressive power for temperature conditioning.

\begin{table}[h]
    \centering
    \caption{Experiments on GridWorld (n=3, H=32). Train with $\beta\sim U[1, 3]$ for Temperature-conditional GFlowNets, GFN are trained with specialized $\beta$.}
    \resizebox{0.9\linewidth}{!}{
    \begin{tabular}{lccccc}
    \toprule
    L1($\times10^{-5}$) & $\beta=1$ & $\beta=2$ & $\beta=3$ & $\beta=4$ & $\beta=5$ \\
    \midrule
    Logit-GFN (Layer-conditioning) & \textbf{3.607 $\pm$ 0.074} & \textbf{2.376 $\pm$ 0.231} & \textbf{1.490 $\pm$ 0.075} & \textbf{1.284 $\pm$ 0.199} & \textbf{1.957 $\pm$ 0.470} \\
    Logit-GFN (Logit only) & 3.926 $\pm$ 0.043 & 2.777 $\pm$ 0.119 & 2.130 $\pm$ 0.158 & 2.426 $\pm$ 0.114 & 2.579 $\pm$ 0.160 \\
    Layer-GFN & 3.621 $\pm$ 0.025 & 2.470 $\pm$ 0.228 & 1.479 $\pm$ 0.193 & 2.926 $\pm$ 0.849 & 5.507 $\pm$ 0.039 \\
    GFN & 3.689 $\pm$ 0.037 & 2.484 $\pm$ 0.071 & 5.363 $\pm$ 0.003 & 5.370 $\pm$ 0.001 & 5.363 $\pm$ 0.001 \\
    \bottomrule
    \end{tabular}}
    \label{tab:gridworld}
\end{table}
\begin{table}[h]
    \centering
    \caption{Experiments on GridWorld (n=4, H=16). Train with $\beta\sim U[1, 3]$ for Temperature-conditional GFlowNets, GFN are trained with specialized $\beta$.}
    \resizebox{0.9\linewidth}{!}{
    \begin{tabular}{lccccc}
    \toprule
    L1($\times10^{-5}$) & $\beta=1$ & $\beta=2$ & $\beta=3$ & $\beta=4$ & $\beta=5$ \\
    \midrule
    Logit-GFN (Layer-conditioning) & \textbf{2.392 $\pm$ 0.020} & \textbf{1.396 $\pm$ 0.033} & \textbf{0.733 $\pm$ 0.047} & \textbf{0.908 $\pm$ 0.011} & 1.226 $\pm$ 0.050 \\
    Logit-GFN (Logit only) & 2.397 $\pm$ 0.017 & 1.463 $\pm$ 0.035 & 1.099 $\pm$ 0.008 & 1.210 $\pm$ 0.046 & \textbf{1.190 $\pm$ 0.010} \\
    Layer-GFN & 2.415 $\pm$ 0.020 & 1.416 $\pm$ 0.033 & 0.840 $\pm$ 0.062 & 1.644 $\pm$ 0.035 & 2.902 $\pm$ 0.018 \\
    GFN & 2.426 $\pm$ 0.015 & 1.403 $\pm$ 0.011 & 0.812 $\pm$ 0.022 & 3.060 $\pm$ 0.001 & 3.075 $\pm$ 0.000 \\
    \bottomrule
    \end{tabular}}
    \label{tab:gridworld2}
\end{table}
\clearpage

\subsection{Experiments on offline generalization}
\begin{figure}[h]
    \centering
    \includegraphics[width=\textwidth]{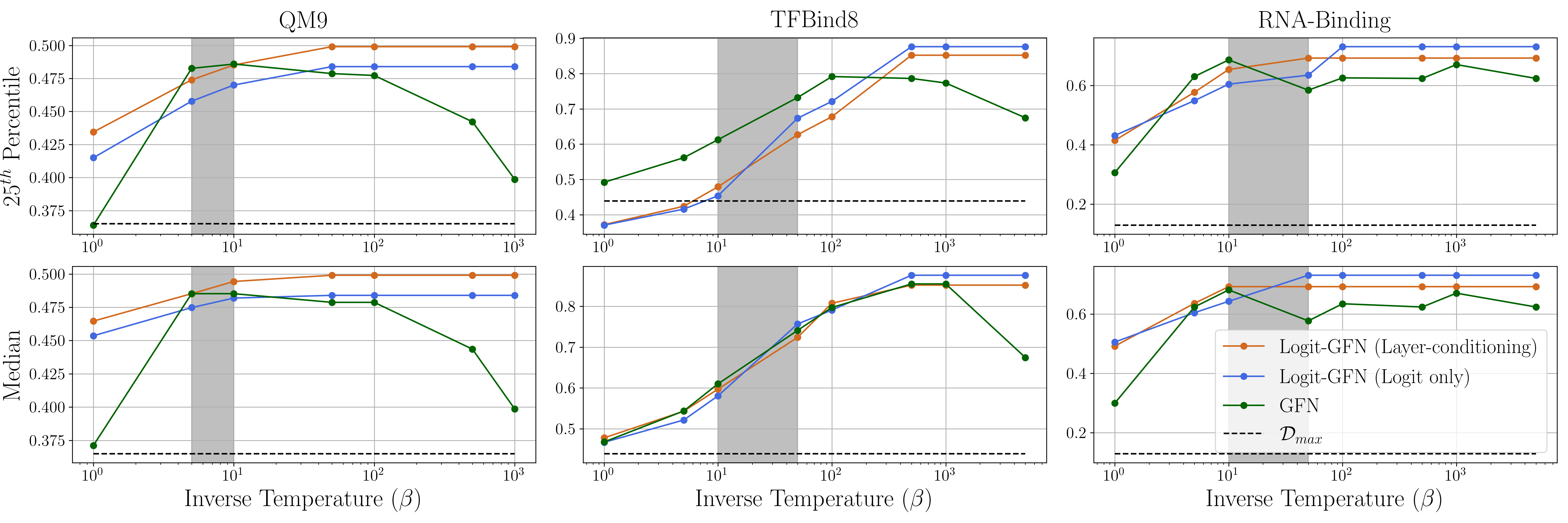}
    \caption{Ablation on Layer-conditioning with Logit Scaling in offline generalization.}\label{fig:concat_ablation}
\end{figure}

\Cref{fig:concat_ablation} visualizes the results of Logit-GFN with layer-conditioning in offline generalization tasks. As depicted in the figure, Logit-GFN with layer-conditioning performs similarly or occasionally better than Logit-GFN without layer-conditioning. In certain scenarios, incorporating a concatenated path can enhance the performance of Logit-GFN. This improvement can be attributed to the simplicity and ease of training of the logit-only method, which results in good performance despite its limited expressiveness compared to Logit-GFN with layer-conditioning.

\subsection{Experiments on online mode-seeking}
\begin{figure}[h]
\begin{minipage}[t]{\textwidth}
    \begin{subfigure}[t]{0.5\textwidth}
        \centering
        \includegraphics[width=\textwidth]{figures/qm9str_modes_div_threshold_05_logit+layer_ablation.png}
        % \caption{Ablation on different $\beta$ for unconditional GFlowNets}
        \label{fig:qm9_logit+layer_ablation_copy}
    \end{subfigure}
    \begin{subfigure}[t]{0.5\textwidth}
        \centering
        \includegraphics[width=\textwidth]
        {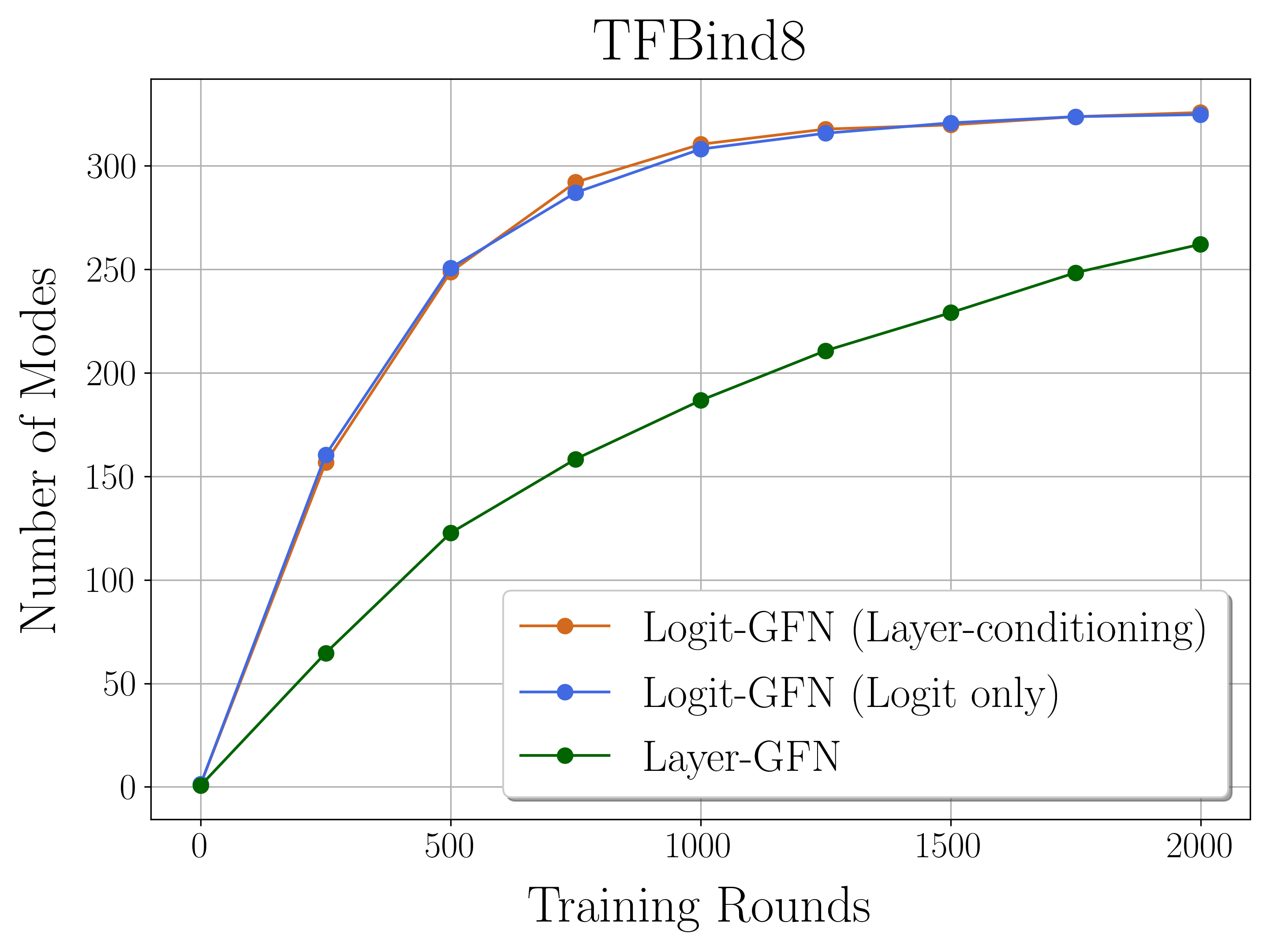}
        \label{fig:tfbind8_logit+layer_ablation}
        % \caption{Ablation on $\beta$ of Unconditional GFlowNets}
    \end{subfigure}
\end{minipage}
\caption{Ablation on Layer-conditioning with Logit Scaling in online mode seeking problems.}\label{fig:logit+layer_ablation}
\end{figure}

We also conducted experiments to evaluate the effect of layer conditioning on Logit-GFN in online mode-seeking problems. \Cref{fig:logit+layer_ablation} illustrates the number of modes discovered during training when applying layer-conditioning to Logit-GFN in the QM9 and TFBind8 tasks. As shown in the figure, while layer-conditioning has a marginal impact on the performance of Logit-GFN, it still outperforms Layer-GFN, demonstrating the superiority of logit scaling.

\clearpage

\section{Online Discovery Algorithm}
\subsection{Algorithm PseudoCode}
% \begin{figure}[t]
% \centering
% \begin{minipage}[t!]{0.9\textwidth}
% \centering
\begin{algorithm}[H]
\small
% \begin{spacing}{1.2}
\caption{Scientific Discovery with Temperature-Conditional GFlowNets}
\label{alg:tempgfn}
\small
\begin{algorithmic}[1]
\State 
Set $\mathcal{D} \gets \emptyset$ \Comment{{\it Initialize dataset.}}
\For{$t=1,\ldots, T$} \Comment{{\it Training $T$ rounds}}
    \State $\beta_1,\ldots,\beta_M \sim P_{\text{exp}}(\beta)$ \Comment{{\it Sample temperatures from exploration query prior.}}
    \For{$m=1,\ldots,M$}
    \State $\tau_m \sim P_F(\tau|\beta = \beta_m;\theta)$ \Comment{{\it Sample trajectories from Logit-GFN.}}
    \State $\mathcal{D} \gets \mathcal{D} \cup \{\tau_m\}$ 
    \EndFor
    \For{$k = 1, \ldots K$} \Comment{{\it Training $K$ epochs per each training rounds}}
        \State Use ADAM for gradually minimizing $\mathcal{L}(\theta;\mathcal{D})$.
    \EndFor
\EndFor
\State Output: $\mathcal{D}$
\end{algorithmic}
% \end{spacing}
\end{algorithm}
% \end{minipage}
% \end{figure}
We present the pseudocode of our online discovery algorithm in \Cref{alg:tempgfn}. For each training round $t$, we sample $M$ (inverse) temperatures from the query prior, $P_{\text{exp}}(\beta)$. Then, we generate trajectories using the policy $P_{F}(\cdot\vert\beta;\theta)$ conditioned on the sampled $\beta$ values. Finally, we iterate the procedure by updating the policy parameters using the TB loss for $K$ epochs per training round.

\subsection{Ablation on different temperature sampling distribution}\label{app:query_distribution}
In this section, we present experiments conducted on different temperature sampling distributions, $P_{\text{exp}}(\beta)$, which is a crucial component for balancing exploration and exploitation in online mode seeking problems. While we use a uniform distribution as a default setting, there are various options available. We present extensive experiment results on various distributions we have tried in this research.

\subsubsection{Stationary distributions}
We first implement various stationary distributions. Among various options, we choose widely used distributions for evaluation as listed below:

\begin{itemize}
    \item \textbf{Uniform}: This is a default setting of our algorithm. We sample $\beta$ from the uniform distribution, i.e, $\beta\sim U^{[a, b]}$.
    \item \textbf{LogUniform}: We sample $\beta$ from the loguniform distribution, which leads to sample high values of $\beta$ with more probability, i.e, $\beta\sim \log({U^{[e^a, e^b]}})$.
    \item \textbf{ExpUniform}: We sample $\beta$ from the expuniform distribution, which leads to sample low values of $\beta$ with more probability, i.e, $\beta\sim e^{U^{[\log a, \log b]}}$.
    \item \textbf{Normal}: We sample $\beta$ from the normal distribution, i.e, $\beta\sim\mathcal{N}(\mu, \sigma^2)$.
\end{itemize}

\begin{figure}[h]
    \centering
    \includegraphics[width=0.9\textwidth]{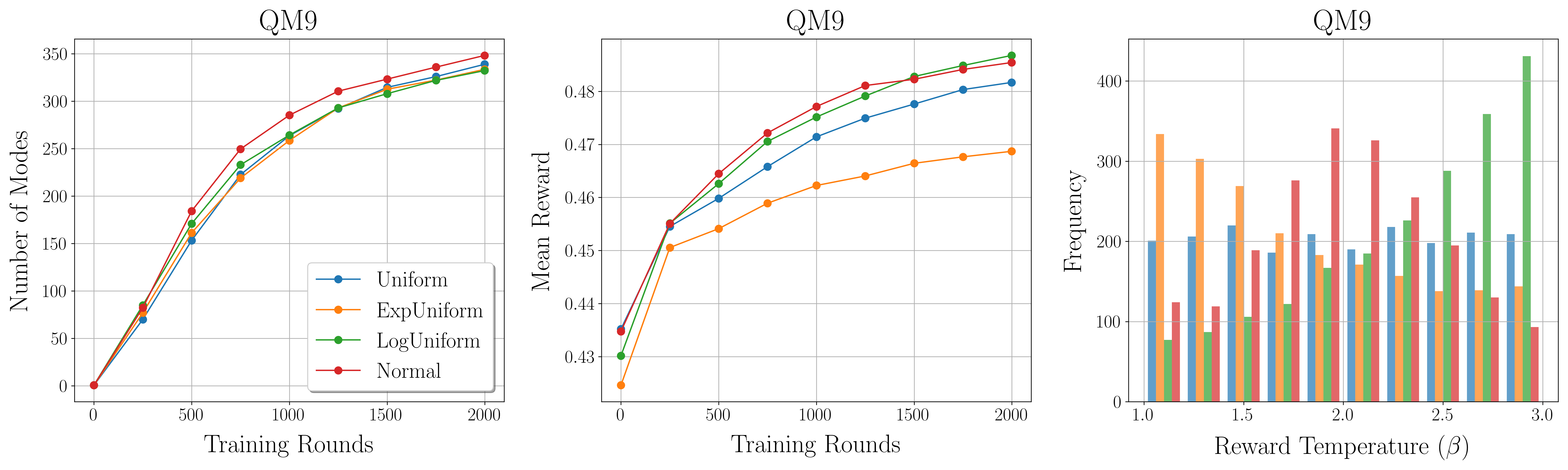}
    
    \vspace{-10pt}
    \caption{Ablation on different temperature sampling distribution. (LogUniform, ExpUniform, Normal)}\label{fig:qm9_query_dist_ablation_add_mean}
\end{figure}

\Cref{fig:qm9_query_dist_ablation_add_mean} (Left) shows the number of modes discovered during training with different query distributions. As shown in the figure, there is no big difference between uniform, lognormal and expuniform distribution. Normal distribution shows slightly better results than other distributions.

In the \Cref{fig:qm9_query_dist_ablation_add_mean} (Middle), we present the average reward computed by samples generated from the policy every 250 training rounds. We find that samples generated from LogUniform and Normal distribution generally give high rewards compared to a naive uniform distribution. Conversely, samples generated from ExpUniform distribution are placed in relatively low score regions. It tells us that our Logit-GFN can generate appropriate samples according to the conditioning variable, $\beta$. \Cref{fig:qm9_query_dist_ablation_add_mean} (Right) shows the histogram of generated $\beta$ values for ease of understanding.
\clearpage

\subsubsection{Dynamic distributions}
We also apply dynamic temperature sampling distributions, which change the distribution across training. Among various options, we implement simulated annealing, one of the most widely used heuristics in combinatorial optimization. We prepare two variants of simulated annealing strategies as follows:

\begin{itemize}
    \item \textbf{Simulated Annealing}: We first define uniform distribution with range 0.5, and shift the mean across the training rounds $t\in[1, T]$ from the lower bound to the upper bound, i.e, $\mu=(b - a)\times(t/T)+a, \: \beta\sim U^{[\text{max}(a, \mu-0.5), \text{min}(\mu+0.5, b)]}$.
    \item \textbf{Simulated Annealing (Inverse)}: In the inverse setting, we shift the mean across the training rounds from the upper bound to the lower bound, i.e, $\mu=(b - a)\times(1 - t/T)+a, \: \beta\sim U^{[\text{max}(a, \mu-0.5), \text{min}(\mu+0.5, b)]}$.
\end{itemize}

\begin{figure}[h]
    \centering
    \includegraphics[width=0.9\textwidth]{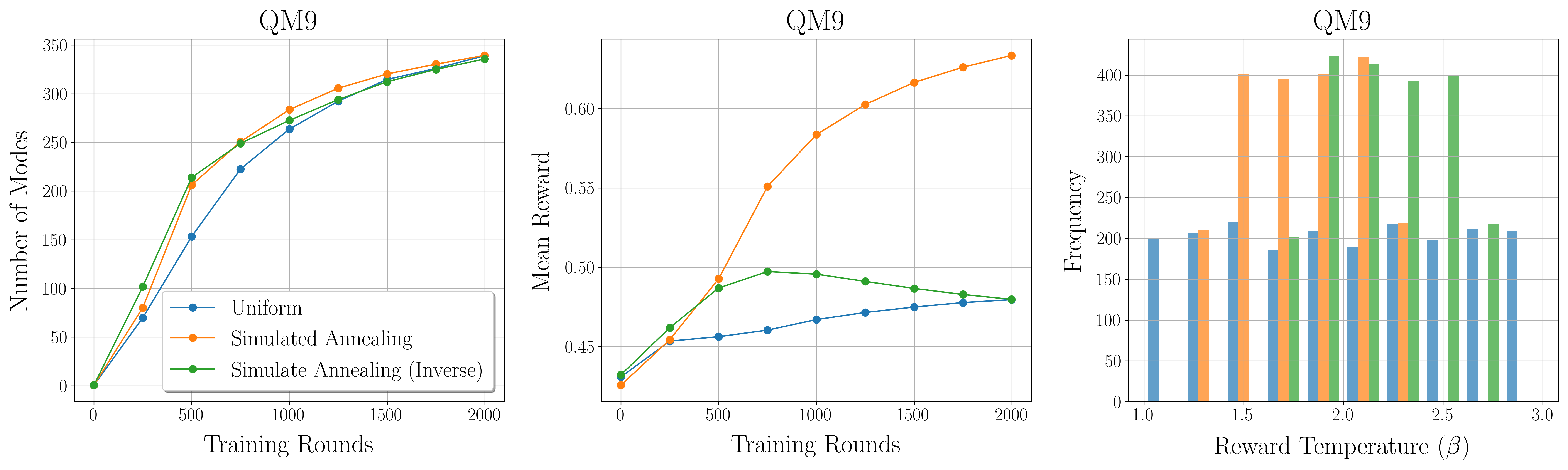}
    \vspace{-10pt}
    \caption{Ablation on different temperature sampling distribution. (Simulated Annealing)}\label{fig:qm9_query_dist_ablation_add_mean_sa}
\end{figure}

As shown in the \Cref{fig:qm9_query_dist_ablation_add_mean_sa}, Simulated annealing demonstrates superior performance compared to a naive uniform distribution. While the inverse approach also exhibits fast convergence at the early stage, it encounters a significant decrease in efficiency as the rounds advance.

We also visualize the average reward of generated samples from the policy at every 250 training rounds in \Cref{fig:qm9_query_dist_ablation_add_mean_sa}. We observe that simulated annealing makes the average reward of samples drastically towards high reward regions. It seems exploration with low $\beta$ at the early stage and focusing on exploitation with high $\beta$ at the later stage can boost the performance in online mode seeking problems. \Cref{fig:qm9_query_dist_ablation_add_mean_sa} (Right) shows the histogram of generated $\beta$ values at training round $t=800$ for ease of understanding.

\end{document}